%% file: main.tex
\RequirePackage{fix-cm}
\documentclass[11pt]{article}
\usepackage{acl}
\usepackage{times,latexsym}
\usepackage[T1]{fontenc}
\usepackage[english]{babel}
\usepackage[utf8]{inputenc}
\usepackage{microtype,inconsolata}
\usepackage{amsmath,amssymb}
\usepackage{booktabs,tabularx,makecell,multirow,array}
\usepackage{tablefootnote}
\usepackage{graphicx,tikz}
\usepackage{pifont}
\usepackage{cleveref}
\usetikzlibrary{arrows.meta}
\usepackage[most]{tcolorbox}

\newtcolorbox{grayquote}{
    width=0.95\linewidth,
    center,
    enhanced,
    frame hidden,
    colback=white,
    coltext=black!80,
    borderline west={1pt}{0pt}{black!40},
    sharp corners,
    boxsep=0pt,
    left=8pt,
    right=0pt,
    top=0pt,
    bottom=0pt,
    before skip=5pt,
    after skip=5pt
}

\definecolor{ORSS}{RGB}{255,233,172}
\definecolor{stopcond}{RGB}{78,150,219}

\title{ReBeCA: Causality-Inspired Behavior Analysis for Understanding the Self-Reflection of Language Models}

\author{Tianqiang Yan$^1$, Lizhen Qu$^1$, Sihan Shang$^2$, Yuheng Li$^{3}$, Song Qiu$^{4}$, Hao Peng$^1$ \\
\textbf{Wenjian Luo$^2$, Jue Xie$^1$, Yuan Gao$^{5,\,}$\thanks{$^*$Corresponding Author}} \\
$^1$\small{\tt firstname.lastname@monash.edu}, $^2$\small{\tt shangsihan@stu.hit.edu.cn, luowenjian@hit.edu.cn} \\
$^3$\small{\tt liyuheng@hainanu.edu.cn}, $^4$\small{\tt 202230242087@mail.scut.edu.cn}, $^5$\small{\tt gaoyuan@cuhk.edu.cn} \\
$^1$Monash University, $^2$Harbin Institute of Technology, Shenzhen \\
$^3$Hainan University, $^4$South China University of Technology \\
$^5$The Chinese University of Hong Kong, Shenzhen
}

\begin{document}
\maketitle
\begin{abstract}
Self-reflection is a basic self-improve\-ment skill of language agents, yet it does not reliably improve their outputs, where the underlying behavioral mechanism remains opaque. We introduce \textbf{ReBeCA} (self-\textbf{Re}flection \textbf{Be}havior explained through \textbf{C}ausality-inspired \textbf{A}nalysis), a framework for identifying and validating outcome-relevant reflective behaviors from recorded trajectories. Re\-BeCA uses \textbf{reflective codes}, an in\-ductive coding schema, to represent each trajectory as a sequence of reflective behaviors. It learns dependencies among these behaviors and the outcome with causal discovery, selects the behaviors that best predict the outcome of held-out trajectories, and tests them for consistency across data subsets and by prompt-based steering. In a controlled case study of four Qwen3 models under fixed-round Self-Refine on mathematical reasoning and translation, ReBeCA identifies compact, round-specific behavioral sets with superior held-out predictive fit and uses prompt-based steering on unseen problems to reveal how individual and combined behavioral directives affect performance (Holm-adjusted $p = .020$ on Qwen3-4B). 
ReBeCA offers a reusable framework that turns agent trajectories into interpretable, verifiable behavioral hypotheses without requiring model internals or preselecting the behaviors to test.
\end{abstract}

\section{Introduction}
\input{Fig/overview}
Language models (LMs) now support iterative agentic workflows~\cite{madaan2023self,shinn2023reflexion}, yet they still hallucinate~\cite{alkaissi2023artificial,zheng2023does}. Self-reflection, in which an LM critiques and revises its own output over several rounds, has been proposed to mitigate such errors~\cite{madaan2023self,yan2025entrospect}. However, growing empirical evidence indicates that self-reflection is not universally effective and may even degrade performance~\cite{huang2023large}. These trajectories support \textit{functional analysis}~\cite{cooper2007applied}: relating intermediate behaviors to outcomes and testing how changes to the reflection process affect performance.

Prior work has documented failure modes of self-reflection and proposed remedies~\cite{kamoi2024can,gou2023critic,chen2023teaching}. Translating these findings into improved reflection requires identifying which behaviors to target and when. However, existing analyses typically focus on predefined factors or internal states~\cite{li2024hindsight,xu2024pride,liu2024intrinsic}, leaving broader trajectory behaviors underexplored. Furthermore, behaviors are coupled across rounds: an apparent correlate of success may merely co-occur with another informative behavior. Prioritizing targets based on such unconditioned associations risks misguiding prompt modifications, particularly under joint elicitation. Researchers therefore need a systematic approach to capture behavioral dependencies from trajectories and empirically test whether prompting them alters outcomes.

We propose ReBeCA (self-\textbf{Re}flection \textbf{Be}havior explained through \textbf{C}ausality-inspired \textbf{A}nalysis), a framework for identifying and verifying behavioral targets from self-reflection trajectories. Figure~\ref{fig:first} outlines its workflow: ReBeCA connects success and failure to interpretable, round-specific behavior combinations and tests how prompts targeting them change outcomes.

\input{Fig/code}

To operationalize reflective behaviors, we introduce \textbf{reflective codes}: a coding schema in which each code names an abstract behavior and describes the text it applies to~\cite{zhong2025hicode} (Figure~\ref{fig:semantic-patterns}). The codes are enumerated by an LM from recorded trajectories and consolidated by humans, providing a candidate pool grounded in observed trajectories and explicit behavioral labels for dependency analysis~\cite{opitz2025interpretable}. GPT-5 annotations agree with human majority labels on $89\%$ of 100 sampled instances (Section~\ref{sec:struct}).

Causal discovery (CD) is the core instrument of ReBeCA. It jointly models conditional dependencies among behaviors and the outcome, using round order to constrain the graph and identify compact outcome-adjacent sets~\cite{chickering2002optimal,pellet2008using}. Prior analyses have examined selected factors or aggregate indicators, such as self-checking accuracy, task verifiability, and reasoning frequency~\cite{wang2024theoretical,zhang2025understanding,stav2026does}.
In iterative self-reflection, however, revisions build on earlier responses and feedback~\cite{madaan2023self,shinn2023reflexion}. A behavior's association with the outcome may arise through its dependency on another behavior: let $C$ and $R$ denote \textit{checking} and \textit{revision}, and $\hat{y}$ the \textit{outcome}. If the underlying dependency is $C \to R \to \hat{y}$, both $C$ and $R$ can be associated with $\hat{y}$ when examined separately, yet $C$ provides no additional information about $\hat{y}$ once $R$ is known, i.e. $C \perp\!\!\!\perp \hat{y} \mid R$. CD extends this check to all behaviors by learning a dependency graph over behaviors and the outcome~\cite{chickering2002optimal}. Dependencies learned from one sample of trajectories may not hold in another, so ReBeCA combines CD with a selection procedure in the spirit of stability selection~\cite{meinshausen2010stability}: it repeats CD on different folds of trajectories and with different score functions, and retains the candidate parent set that best predicts the outcome on held-out trajectories. From this set, ReBeCA derives behavioral hypotheses specific to each task and reflection round (see Stage \ding{184} in Figure~\ref{fig:first}).

In our controlled case study, frozen models generate fixed-round trajectories without external tools or human feedback, and Consistency-Enhanced Self-Refine (CESR; Section~\ref{sec:cesr}) selects a representative output at each step~\cite{wang2023self,kuhn2023semantic,laban2025llms}.

We evaluate ReBeCA across four Qwen3 models on mathematical reasoning and translation tasks. Out of 30 candidate variables, ReBeCA selects one or two behaviors per task at specific rounds (e.g., \textit{Clarity and Organization} at round 2 and \textit{Specificity} at round 4 on MATH). These sparse sets of behaviors predict the outcomes of held-out trajectories more accurately than the full set of variables combined; the learned graphs organize broader associations into temporal persistence, within-round co-occurrence, and outcome-adjacent dependencies. Furthermore, prompt-based elicitation of the selected behaviors on unseen problems exposes substantial performance differences between individual and combined steering. Our primary contribution lies beyond these specific behavioral findings: by integrating interpretable coding, structure learning, and validation, ReBeCA formalizes how critical agentic behaviors are identified from observed, unstructured trajectories and empirically evaluated. This provides a quantitative basis for understanding modern agents through the relationship between their behaviors and task outcomes.

\section{Related Work}
\label{sec:rw}

\subsection{Understanding Self-Reflection}
Self-reflection prompts language models to critique and revise their own outputs, and has been adopted to reduce errors and support agentic behaviors~\cite{madaan2023self,shinn2023reflexion}. Subsequent studies show that it is not universally reliable: performance can degrade on certain tasks~\cite{li2024hindsight}, and some earlier gains relied on oracle labels~\cite{huang2023large,kim2023language}. Proposed remedies add external tools, execution feedback, or trained critics~\cite{gou2023critic,chen2023teaching,paul2023refiner,wang2023shepherd}, and a survey finds that self-correction succeeds mainly when reliable external feedback is available~\cite{kamoi2024can}. These remedies improve performance in specific settings, motivating systematic analysis of the intermediate behaviors they target.

Analyses of why self-reflection succeeds or fails remain limited in two ways. Black-box analyses relate the outcome to factors selected in advance, such as self-bias~\cite{xu2024pride}, uncertainty~\cite{kirchhof2025self}, or modify the reflection procedure to test specific capabilities~\cite{li2024hindsight}. These studies test specified explanatory factors; ReBeCA complements them by deriving behavioral candidates from trajectories and analyzing their dependencies. White-box analyses inspect hidden states during self-correction~\cite{liu2024intrinsic}, but their findings are difficult to translate into behaviors that can be observed or steered. ReBeCA selects round-specific behavioral sets by held-out prediction and evaluates their prompt-based elicitation.

\subsection{Temporal Process Modeling}
\label{rw:tpm}
Self-reflection trajectories can be modeled as temporal processes in several ways, each answering a different question. Predictive sequence models estimate the outcome from the whole trajectory and can be paired with feature attributions, but accounting for dependent inputs is essential for interpreting such attributions~\cite{aas2021explaining}. Temporal causal discovery offers methods for diverse sequential settings~\cite{runge2018chaos,gong2024causal}. Here, self-reflection yields many short trajectories with shared protocol positions, making round-indexed variables a natural representation. ReBeCA therefore treats each behavior at each round as a separate variable, encodes the round order as tiered background knowledge~\cite{andrews2020completeness}, and applies score-based CD across trajectories~\cite{chickering2002optimal}. Predictive accuracy alone does not determine which behaviors are useful to steer~\cite{shmueli2010explain,pearl2009causality}. ReBeCA uses conditional predictive relevance and consistency across data subsets~\cite{meinshausen2010stability} to prioritize targets for prompt-based tests.

\section{Problem Formulation}
\label{sec:prob}
Iterative self-reflection generates natural-language trajectories~\cite{madaan2023self,shinn2023reflexion}, adding intricacies as key signals are submerged by unstructured, redundant information. However, reflective behaviors remain observable and steerable~\cite{miehling2025evaluating}. We therefore study self-reflection through the lens of agentic behavioral analysis~\cite{rahwan2019machine}, framing it as a discovery problem: \textit{hypothesizing which round-specific behaviors the outcome depends on, and testing these hypotheses empirically}~\cite{langley1987scientific}.

\noindent\textbf{Trajectories and outcome.} 
Given a query $x \in \mathcal{C}$, a model $\mathcal{M}_\theta$ under $T$-round Self-Refine produces a trajectory $\tau = (y_0, f_{b_0}, y_1, \dots, f_{b_{T-1}}, y_T)$ of responses and feedback~\cite{madaan2023self} (Appendix~\ref{sec:isr}). The task outcome is defined as a discrete variable $\hat{y} = g(x, y_0, y_T)$.

\input{Fig/pipeline}

\noindent\textbf{Behavioral variables.} 
A codebook of $N$ reflective codes (Section~\ref{sec:struct}) discretizes each trajectory into a binary matrix $\mathcal{S} \in \{0,1\}^{T \times N}$, where $\mathcal{S}_{t,n} = 1$ indicates the presence of code $n$ at round $t$. Together with optional round-level indicators, these yield the variable set $\Omega$. Round indices align shared protocol positions across queries: each iteration applies the same feedback and refinement prompts. Code activation therefore records variation in generated behavior under a fixed prompting protocol. Temporally, variables can only depend on those from the same or preceding rounds, while $\hat{y}$ temporally succeeds all rounds.

\noindent\textbf{Discovery target.} 
We seek the minimal set of variables conditioning on which renders the outcome independent of all others:
\begin{equation}
\mathrm{Pa}_{\hat{y}} = \mathop{\arg\min}_{P \subseteq \Omega} |P| \,\, \text{s.t.} \,\, \hat{y} \perp\!\!\!\perp \Omega \setminus P \mid P . \label{eq:target}
\end{equation}

This set forms the Markov boundary of $\hat{y}$, unique when the distribution satisfies the intersection property (e.g., under strict positivity)~\cite{statnikov2013algorithms}. If the distribution is further Markov and faithful to a DAG over $\Omega \cup \{\hat{y}\}$ respecting round order, $\hat{y}$ has no children as a terminal outcome. Its Markov boundary thus reduces to its graphical parent set $\mathrm{Pa}_{\hat{y}}$~\cite{pellet2008using}, which motivates selecting outcome-adjacent variables from learned graphs. We term its members \textit{direct dependencies}. Since unobserved confounders may exist, these dependencies need not be true causal parents; we treat them as candidate hypotheses and validate them via prompt steering (C4).

\noindent\textbf{Evaluation criteria.}
In finite-sample regimes, Eq.~\eqref{eq:target} must be estimated empirically. Inspired by CD, stability selection~\cite{meinshausen2010stability}, and prompt steering strategies~\cite{miehling2025evaluating}, ReBeCA evaluates candidate dependencies against four criteria (Figure~\ref{fig:validation}), each addressing a distinct empirical question:

\noindent\textit{(C1) Stable selection}: selects $\hat{P}$ maximizing held-out CVLL across folds and scoring functions (Stage~1), comparing candidates generated under different folds and scores;

\noindent\textit{(C2) Boundary comparison}: verifies that $\hat{P}$ outperforms both the null ($\emptyset$) and full ($\Omega$) models under the same held-out metric (Stage~2), assessing predictive value and compactness;

\noindent\textit{(C3) Consistent strength}: assesses predictive strength across data partitions (Stage~3), quantifying variation across subsets;

\noindent\textit{(C4) Steerability}: tests whether actively eliciting $\hat{P}$ alters performance on unseen problems (Section~\ref{sec:intervention}), guarding against observational associations that fail under steering.

All findings are conditional on the model $\mathcal{M}_\theta$, the $T$-round Self-Refine setup, and the considered reflective codes.

\section{The ReBeCA Pipeline}
\input{Fig/cesr}
\subsection{Collecting Representative Trajectories}
\label{sec:cesr}
Single-sample trajectories can compound sampling variation across rounds~\cite{shorinwa2025survey}, whereas exploring all sampled branches requires exponentially many generations in $T$. Consistency-Enhanced Self-Refine (CESR; Figure~\ref{fig:second}) follows a representative generation at each step, using a fixed sample budget per step and a total budget linear in $T$. At each step, CESR draws $K$ samples, embeds them via OpenAI's \texttt{text-embedding-3-large}, and clusters them with HDBSCAN~\cite{mcinnes2017hdbscan} under cosine distance, grouping semantically similar generations in embedding space~\cite{kuhn2023semantic}. The medoid of the largest cluster, i.e., its most central and representative sample, is selected as the output of that turn and conditions the subsequent step. This formulation extends the majority-voting principle of self-consistency~\cite{wang2023self} to free-form generations that rarely match verbatim: the largest sampled cluster identifies recurring content, and its medoid supplies a representative, authentic generation. For our study, we set $T = 5$, $K = 20$, temperature $0.7$, and top-p $0.8$, yielding one representative trajectory per query and model. 

\input{Fig/graphs}

\subsection{Coding Reflective Behaviors}
\label{sec:struct}
Reflective codes map unstructured natural-language trajectories into the binary variable space $\Omega$ defined in Section~\ref{sec:prob}.
Reflective codes give each variable an explicit behavioral meaning and limit the number of variables for score-based graph search~\cite{zanga2022survey}.

We construct the behavioral codebook in two stages:
\begin{enumerate}
\item \textit{Candidate induction}: GPT-5 induces candidate behaviors from a random subset of trajectories, blinded to their final outcomes.
\item \textit{Consolidation and filtering}: We prune rare behaviors occurring in fewer than $1\%$ of rounds to avoid unstable dependency estimates~\cite{chiu2023towards}, and merge semantically overlapping codes.
\end{enumerate}

The consolidated definitions are detailed in Appendix~\ref{sec:sp}. Using this finalized codebook, GPT-5 annotates all rounds across the trajectories. GPT-5 agreed with the majority label of three independent human evaluators on 89 of 100 randomly sampled instances ($89\%$).

\subsection{Learning and Validating Direct Dependencies}
\label{sec:icp}
This procedure estimates the target dependency set in Eq.~\eqref{eq:target} and evaluates it against criteria C1--C3; criterion C4 is evaluated by prompt-based steering in Section~\ref{sec:intervention}.

\noindent\textbf{Stage 1: Stability-based selection (C1).} We learn dependency graphs with Greedy Equivalence Search (GES)~\cite{chickering2002optimal} in pgmpy~\cite{ankan2024pgmpy} (Appendix~\ref{sec:ges}). The search scores candidate graphs by balancing data fit and model complexity~\cite{hastie2009elements}.
We encode $\hat{y}$ as a terminal node by forbidding its outgoing edges. Consequently, every variable adjacent to $\hat{y}$ in a learned graph is a candidate graphical parent.
Learned dependency structures  can vary with the data sample and scoring objective~\cite{scutari2016empirical}; we therefore run GES on each of $F$ training folds under four scores: AIC~\cite{akaike1974new}, BIC~\cite{schwarz1978estimating}, BDeu~\cite{heckerman1995learning}, and BDs~\cite{scutari2016empirical}. Each run yields a candidate parent set. We select the candidate that maximizes cross-validated log-likelihood (CVLL): conditional probabilities are fitted on each training fold and used to score held-out outcomes, with mean log probabilities averaged across folds (Appendix~\ref{sec:bds}). 

\noindent\textbf{Stage 2: Boundary comparison (C2).} GES explores a restricted path within the $2^{\vert{}\Omega\vert{}}$ combinatorial candidate space. We benchmark the selected set $\hat{P}$ against two boundary extremes under the identical held-out CVLL metric: the null model ($\emptyset$), which predicts $\hat{y}$ without behavioral variables, and the full model ($\Omega$), which incorporates all available behaviors. Outperforming the null model establishes the predictive value of $\hat{P}$; outperforming the full model shows that this compact behavioral representation achieves better held-out predictive fit than exhaustive feature inclusion. 

\input{Tab/selection}

\noindent\textbf{Stage 3: Consistency across folds (C3).} A high average CVLL does not guarantee that each selected behavior maintains a stable predictive relationship with $\hat{y}$ across data splits. For a binary outcome, we fit a linear probability model of $\hat{y}$ on $\hat{P}$ within each of $M$ disjoint folds, partition the folds into groups, and test whether regression coefficients, including the intercept, differ across groups in mean (one-way ANOVA~\cite{fisher1930statistical}) or variance (Levene's test~\cite{levene1960robust}). With binary behavioral predictors and small fold sizes, a linear model keeps coefficients directly interpretable as conditional differences in success rate and avoids separation issues common to logistic regression. The empirical consistency criterion is satisfied when neither test reaches significance ($p > .05$) and between-group effect sizes remain small ($\eta^2_\mathrm{p} < .06$). This procedure adapts the hypothesis-testing formulation of invariant prediction to evaluate the empirical consistency of predictive strength across random partitions rather than disparate causal environments~\cite{peters2016causal}. For a multilevel outcome, we report the empirical conditional distribution of $\hat{y}$ given $\hat{P}$ instead (Section~\ref{sec:obs}).

\section{Case Study on the Qwen3 Model Family}
\label{sec:exp}
We evaluate ReBeCA on four Qwen3 models (0.6B to 8B) to study behavioral variation across model sizes under Self-Refine with CESR sampling~\cite{yang2025qwen3,madaan2023self}. We examine two tasks (Appendix~\ref{sec:study_config}): MATH~\cite{hendrycks2021math}, with binary outcome $\hat{y}$ indicating answer correctness ($y_T$); and BOUQuET translation into English~\cite{andrews2025bouquet}, with ternary $\hat{y}$ capturing metric change (higher, equal, or lower quality relative to $y_0$) under COMET or MetricX~\cite{rei2020comet,juraska2023metricx}. Pooling across models yields 417 MATH and 400 BOUQuET trajectories. Stages 1--2 hold out one MATH subject or BOUQuET source language per fold, grouping all model trajectories within that subject or language. Across $T=5$ rounds, the candidate space comprises $\vert{}\Omega\vert{} = 30$ variables, formed by four shared reflective codes, namely Completeness, Specificity (Sp), Clarity and Organization (CO), and Depth of Analysis (DA; Appendix~\ref{sec:sp}), and two round-level uncertainty metrics (MI, PPL; \citealp{yan2025entrospect}) per round (e.g., $\mathrm{CO}_2$ for CO at round 2). Stages 1--2 use $F = 7$ folds on MATH and $F = 5$ on BOUQuET, while Stage~3 uses $M = 20$ random trajectory folds in four groups. 

\subsection{Observational Analysis (C1--C3)}
\label{sec:obs}
Pairwise association alone cannot isolate which behaviors the reflection outcome depends on, as statistical associations are widespread: on BOUQuET, 11 of the 15 active codes associate with the outcome under COMET and 12 under MetricX ($\chi^2$ tests, $p < .05$; Figure~\ref{fig:structure}b). ReBeCA selects $\{\mathrm{CO}_2, \mathrm{Sp}_4\}$ on MATH and $\{\mathrm{CO}_3\}$ on BOUQuET under both metrics. We assess these sets using C1--C3 and examine how the other associated behaviors connect to them. 

\noindent\textbf{(C1) Which behaviors best predict the outcome?}
Among the candidate parent sets returned across $4F$ GES runs, ReBeCA selects the configuration maximizing held-out CVLL (Table~\ref{tab:selection}). On MATH, $\{\mathrm{CO}_2, \mathrm{Sp}_4\}$ ranks first ($-0.434$), ahead of $\{\mathrm{CO}_2\}$ alone ($-0.445$), while candidates incorporating other variables (such as CO at later rounds) score lower. On BOUQuET, the single-variable set $\{\mathrm{CO}_3\}$ achieves the highest CVLL under both metrics; augmenting it with $\mathrm{CO}_4$ (the runner-up candidate) slightly degrades predictive fit ($-0.978$ vs. $-0.981$ under COMET). This comparison identifies round-specific behavioral sets with the strongest held-out predictive fit among the learned candidates.

\noindent\textbf{(C2) How does compact selection compare with exhaustive inclusion?}
We benchmark the selected sets against both the null model ($\emptyset$) and the full model ($\Omega$, 30 variables) on held-out trajectories. The selected sets consistently outperform the null baseline, reducing negative log-likelihood by $22.4\%$ on MATH and $4$--$6\%$ on BOUQuET, while exceeding the full model by $49.6\%$ and $12$--$15\%$, respectively. Strikingly, the full model underperforms even the null model, with held-out CVLL degrading as candidate sets expand (Spearman's $\rho = -0.64$ on MATH, $-0.94$ on BOUQuET). These results demonstrate the predictive advantage of ReBeCA's compact, round-specific behavioral representation over the full-variable model on held-out trajectories.

\input{Tab/stability}

\begin{grayquote}
\textbf{Takeaway (C1--C2).} \textit{ReBeCA condenses 30 candidate variables to just one or two round-specific behaviors per task, which predict held-out outcomes more accurately than all 30 variables combined.}
\end{grayquote}

\noindent\textbf{(C3a) Is linear dependence stable across data folds (MATH)?} We evaluate C3 using the linear probability model of Stage~3 on the binary MATH outcome, reserving the ternary BOUQuET outcome for distributional analysis below. Across the $M$ disjoint folds, between-group differences in coefficient means and variances are small (all $p > .30$, $\eta^2_\mathrm{p} \le .045$; Table~\ref{tab:stability}a). Holding the other behavior fixed, $\mathrm{CO}_2$ and $\mathrm{Sp}_4$ are associated with success rates $29.8$ and $16.0$ percentage points higher, respectively; both 95\% CIs exclude zero. These results support consistent predictive relationships across the sampled partitions. C4 tests prompts that elicit these behaviors.

\noindent\textbf{(C3b) Is distributional dependence consistent across metrics (BOUQuET)?}
For the ternary outcome, we evaluate C3 via the distribution of $\hat{y}$ conditioned on $\mathrm{CO}_3$ (Table~\ref{tab:stability}b). Across both COMET and MetricX, $\mathrm{CO}_3$ exhibits a consistent, marked dependence pattern: equal-score outcomes fall from $38.8\%$ to $8.1\%$ under COMET and from $39.5\%$ to $9.3\%$ under MetricX, with both improvement and degradation becoming more frequent. This agreement across metrics links round-3 CO to changes in translation quality in both directions (Table~\ref{tab:qual}).

\begin{grayquote}
\textbf{Takeaway (C3).} \textit{Discovered dependencies exhibit consistent predictive strength across data folds on MATH, and maintain a consistent dependence pattern across both metrics on BOUQuET.}
\end{grayquote}

\input{Tab/examples}

\noindent\textbf{Why are the other associated behaviors not selected?}
The learned graphs organize associated behaviors into pathways through the selected variables ($C \to R \to \hat{y}$). On BOUQuET, $\mathrm{CO}_2$ associates strongly with the outcome alone ($p < .001$), but among trajectories with identical codes at rounds 3 and 4, the conditional association is not significant (stratified likelihood-ratio tests: $p = .19$ under COMET, $.07$ under MetricX). The learned graph topologies explain this phenomenon (Figure~\ref{fig:structure}): earlier behaviors connect to subsequent selected parents rather than directly to $\hat{y}$. Furthermore, the vast majority of recurring code-to-code edges (31 of 37 on MATH, 28 of 36 on BOUQuET) reflect temporal self-persistence across rounds or co-occurrence within rounds. Rather than treating correlated behaviors as independent predictors, ReBeCA organizes these associations into an interpretable dependency structure centered on the outcome.

\begin{grayquote}
\textbf{Takeaway.} \textit{ReBeCA uncovers the dependency topology of self-reflection trajectories, resolving marginal associations into temporal persistence, intra-round co-occurrence, and direct dependencies.}
\end{grayquote}

\subsection{Steering Experiment (C4)}
\label{sec:intervention}
While C1--C3 validate observational dependencies, C4 tests whether actively steering these nominated behaviors alters self-reflection outcomes on unseen, more challenging problems. In mathematical reasoning, the selected set $\{\mathrm{CO}_2, \mathrm{Sp}_4\}$ enables factorial steering. We benchmark on 50 non-overlapping problems from AIME (an Olympiad-level dataset; \citealp{aime_1983_2024}) under four conditions: an unsteered baseline, and prompts that restrict reflection to the dimension of $\mathrm{CO}_2$, $\mathrm{Sp}_4$, or both (Appendix~\ref{sec:steer}). Evaluating all conditions on the identical 50 problems per model establishes a matched design that strictly controls for problem difficulty and model capacity, tested via Cochran's Q test for matched binary outcomes~\cite{cochran1950comparison}.

Steering modulates success rates on these unseen, harder problems (Table~\ref{tab:optimization_results}). Eliciting either behavior individually increases the overall number of correct answers from 59 to 64 ($\mathrm{CO}_2$) and 62 ($\mathrm{Sp}_4$) out of 200, consistent with the positive coefficients estimated in Stage~3. These gains are concentrated on Qwen3-4B, where success rates differ significantly across conditions (Holm-adjusted $p = .020$, Cochran's Q), rising from 21 to 26 under $\mathrm{CO}_2$ and 33 under $\mathrm{Sp}_4$, whereas Qwen3-0.6B solves at most one problem across all settings. Joint steering yields 49 correct solutions in total, below the unsteered baseline in every model and below the better single-code condition. On Qwen3-4B, $\mathrm{Sp}_4$ alone also outperforms joint steering (33 vs.\ 19; exact McNemar, Holm-adjusted $p = .012$ across 24 pairwise comparisons). This contrast demonstrates why ReBeCA tests prompt combinations explicitly: favorable observational coefficients do not ensure gains when the corresponding prompts are combined.

\input{Tab/steering}

\begin{grayquote} 
\textbf{Takeaway (C4).} 
\textit{On unseen and harder problems, individual steering yields higher aggregate success than joint steering; the strongest paired contrast is $\mathrm{Sp}_4$ versus joint steering on Qwen3-4B. ReBeCA thus connects discovered behaviors to experimentally distinguishable prompt strategies.}
\end{grayquote}

\subsection{Discussion}
\label{sec:synthesis}
Our empirical results yield three actionable insights. First, dependencies are round-specific ($\mathrm{CO}_2$ and $\mathrm{Sp}_4$ on MATH; $\mathrm{CO}_3$ on BOUQuET): within a fixed-round protocol, when a behavior occurs matters, not only whether it occurs. Second, behaviors serve distinct domain functions: CO and Sp distinguish more successful reasoning trajectories ($+16.0$ to $+29.8$ conditional percentage points), while CO in translation accompanies more changes in quality in both directions (Table~\ref{tab:stability}b). Third, the gap between observational co-occurrence and joint steering performance on AIME ($62$--$64$ single vs.\ $49$ joint) shows why prompt combinations require direct testing. ReBeCA connects behavioral discovery to these tests, identifying both promising individual strategies and combinations that reduce performance.

\section{Conclusion}
ReBeCA identifies which behaviors in recorded agent trajectories warrant testing and evaluates prompts that elicit them at specific rounds. We identify sparse, round-specific behavioral dependencies with strong held-out predictive fit and show how prompts targeting them alter performance, including losses under joint steering. ReBeCA thus supports hypothesis-driven behavioral analysis of language agents, from observations to targeted tests.

\section*{Limitations}
Our empirical case study is restricted to four scales of a single model family (Qwen3) and two tasks (${\sim}400$ trajectories each) under a fixed five-round protocol. Consequently, the discovered dependencies remain conditional on this architectural family, the predefined round budget, the induced codebook, and the pooled trajectory distribution; they may not transfer directly to other model families, prompt templates, or dynamic reflection workflows. Furthermore, treating each iteration as a discrete round index assumes uniform step alignment across trajectories, leaving the modeling of variable-length conversational sequences to future work.

To manage branching complexity, CESR tracks only the medoid of the largest sampled embedding cluster with a fixed budget of $K = 20$ samples per step. While this focuses analysis on consensus behavioral paths, it inherently under-represents low-probability or tail trajectories that may contain distinct failure modes. Additionally, a uniform sample allocation does not adapt dynamically to varying query difficulties or model capacities.

The behavioral codebook relies on LLM-assisted candidate induction followed by manual consolidation. Although this approach ensures human-interpretable categories and bounds feature dimensionality for score-based graph search, it operates within a closed candidate vocabulary. The codebook may omit behaviors outside the induction prompts, and annotation uses reference outputs. Per-code and inter-human reliability, alongside sensitivity to coding choices, remain directions for validation.

The identified behaviors denote direct graphical dependencies under the Markov and faithfulness assumptions of Section~\ref{sec:prob}. If latent confounders simultaneously influence intermediate behaviors and final outcomes, selected codes may partially absorb these unobserved factors. In addition, prompt-based steering cannot guarantee surgical behavioral isolation: natural-language instructions may alter untargeted aspects of generation, such as length or tone, and restricting reflection to one dimension also omits the others at that round, which may contribute to the decline under joint steering alongside instruction interference. CVLL is used for candidate selection and comparison; an outer evaluation would estimate performance independently of that selection. C1--C3 provide empirical evidence for compact dependencies, while exact Markov-boundary recovery remains unverified. C4 currently covers MATH-derived behaviors. Extending steering to translation and broader tasks is future work.

\section*{Ethical Considerations}
This study was conducted in adherence to the ACL Code of Ethics.

\paragraph{Machine Behavior and Responsible Interpretation.}
Applications that use these behavioral findings to guide model outputs should evaluate both improvements and newly introduced errors in their own task setting. Our experiments assess benchmark performance, yet they do not establish the safety of these prompts for high-stakes use.

\paragraph{Human Participation and Quality Checking.}
To validate annotation quality, limited human evaluation was conducted on 100 randomly sampled trajectory instances. The evaluators were qualified academic researchers (including authors and graduate researchers who consented to participate as volunteers without monetary compensation). The evaluation involved solely reading and coding public benchmark outputs and mathematical solutions.

\paragraph{Data, Assets, and Environmental Impact.}
All experiments utilize established, publicly available research datasets (MATH, BOUQuET, and AIME) and open-weights models (the Qwen3 family). All assets were accessed and employed in strict accordance with their respective open-source licenses and intended research usage. Computational resources and generation counts are reported in Appendix~\ref{sec:study_config}.

\paragraph{Use of AI Assistants.}
We used AI assistants for language polishing, \LaTeX{} formatting, and coding support. The authors reviewed and revised AI-assisted content and take responsibility for the final methods, analyses, claims, and citations.

\section*{Acknowledgements}
We appreciate anonymous reviewers, ACs and the SAC for their time and insightful comments. 

This work was supported in part by the Shenzhen Science and Technology Program under Grant No.~JSGGKQTD20221101115656029; the Shenzhen Institute of Artificial Intelligence and Robotics for Society under Grant Nos.~AC01202601012 and AC01202601005; and the Shenzhen Science and Technology Program under Grant No.~ZDCY20250901104706008.


\bibliography{reference}

\input{appendix}

\end{document}

%% file: Fig/overview.tex
\begin{figure}[!t]
\centering
\includegraphics[width=0.95\linewidth]{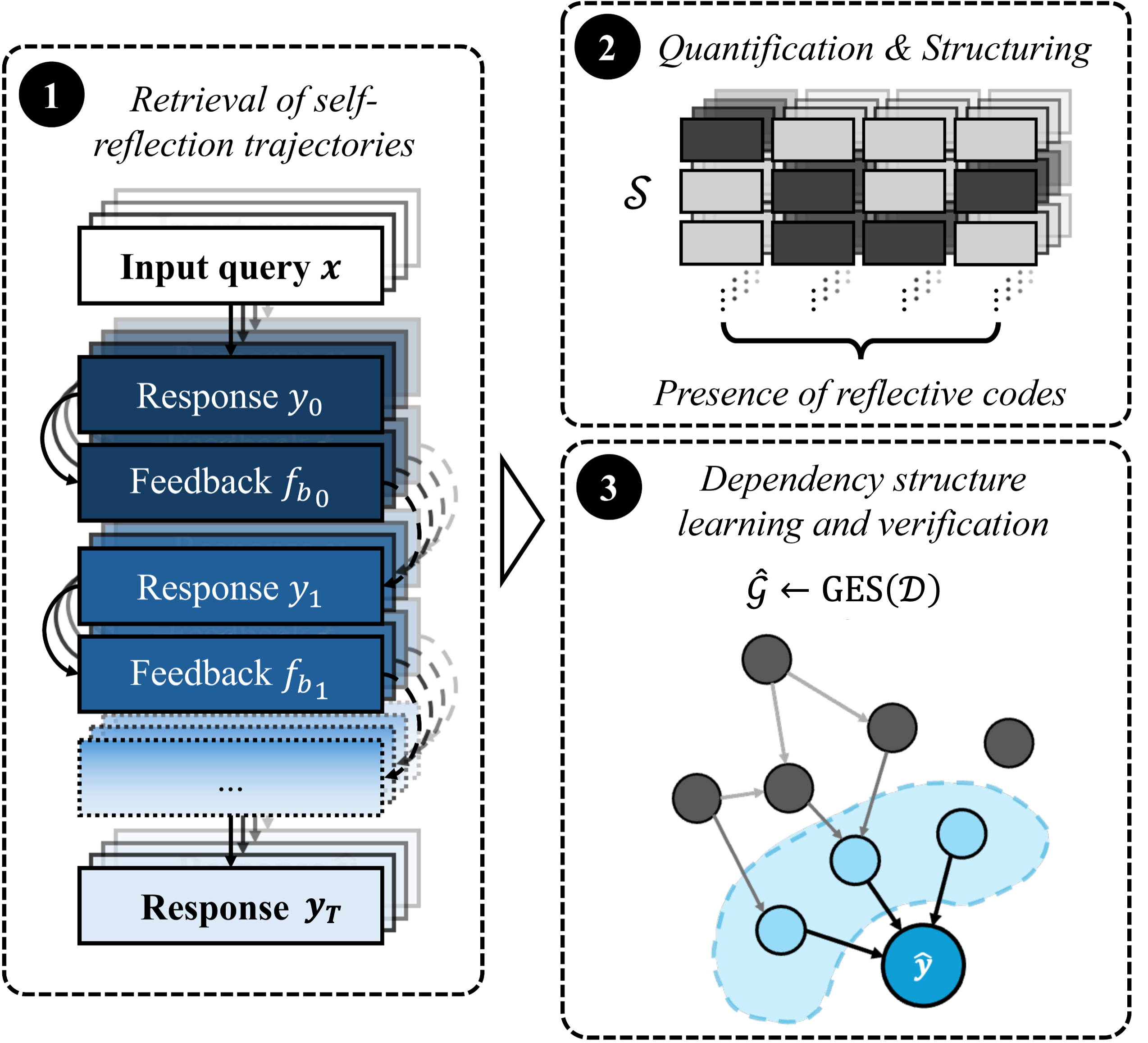}
\caption{Overview of ReBeCA. (1) Self-Refine produces trajectories; (2) reflective codes record which behaviors occur at each round, giving the binary matrix $\mathcal{S}$; (3) causal discovery learns a dependency graph $\hat{\mathcal{G}}$ over the coded behaviors and $\hat{y}$, and the parents of $\hat{y}$ (shaded) are retained as behavioral hypotheses and tested on held-out trajectories and by prompt-based steering.}
\label{fig:first}
\end{figure}

%% file: Fig/code.tex
\begin{figure}[t]
  \centering
  \resizebox{0.95\columnwidth}{!}{%
    \begin{tikzpicture}[
      x=1cm,y=1cm,
      every node/.style={
        font=\normalfont\rmfamily\fontsize{7.5}{8.5}\selectfont,
        inner sep=0pt,text=black
      },
      heading/.style={
        font=\normalfont\rmfamily\fontsize{6.5}{7.5}\selectfont,
        anchor=west
      },
      body/.style={anchor=west},
      card/.style={rounded corners=2pt,line width=0.5pt}
    ]
      \path[use as bounding box] (0,0) rectangle (8.30,1.85);
      \node[anchor=west] at (0,1.69) {\textbf{Reflective code}};
      \node[anchor=west] at (5.75,1.69) {\textbf{Text spans}};
      \begin{scope}
        \clip[rounded corners=2pt] (0.01,0.10) rectangle (5.15,1.50);
        \fill[black] (0.01,0.10) rectangle (2.10,1.50);
        \fill[black!7] (2.10,0.10) rectangle (5.15,1.50);
      \end{scope}
      \draw[card,black] (0.01,0.10) rectangle (5.15,1.50);
      \node[heading,text=white] at (0.60,1.29) {\textbf{Concept}};
      \node[text=white,align=center] at (1.10,0.765)
        {Clarity and\\Organization};
      \node at (2.28,0.77) {$+$};
      \node[heading] at (2.60,1.29) {\textbf{Coding description}};
      \node[body] at (2.60,0.91) {Addresses structure};
      \node[body] at (2.60,0.61) {or readability.};

      \draw[black,line width=0.5pt]
        (5.15,0.80) -- (5.43,0.80);
      \draw[black,line width=0.5pt]
        (5.43,1.19) -- (5.43,0.41);
      \draw[black,line width=0.5pt]
        (5.43,1.19) -- (5.72,1.19);
      \draw[black,line width=0.5pt]
        (5.43,0.41) -- (5.72,0.41);

      \draw[card,draw=black,fill=white]
        (5.75,0.88) rectangle (8.29,1.50);
      \draw[card,draw=black,fill=white]
        (5.75,0.10) rectangle (8.29,0.72);
      \node[body] at (5.93,1.285)
  {\textquotedblleft Reorder these steps};
\node[body] at (5.93,1.035) {for clarity.\textquotedblright};
\node[body] at (5.93,0.505) {\textquotedblleft Make the structure};
\node[body] at (5.93,0.255) { clearer.\textquotedblright};
    \end{tikzpicture}%
  }
  \caption{Illustrative spans for Clarity and Organization. A reflective code pairs a concept with a coding description that specifies which text spans express it; each code is recorded at each round as present or absent.}
  \label{fig:semantic-patterns}
\end{figure}

%% file: Fig/pipeline.tex
\begin{figure}[t]
\centering
\definecolor{dpDark}{RGB}{33,90,160}
\definecolor{dpMid}{RGB}{96,150,205}
\definecolor{dpLight}{RGB}{214,230,246}
\definecolor{dpLine}{RGB}{43,108,176}
\definecolor{dpGrey}{RGB}{238,238,238}
\definecolor{dpGreyLine}{RGB}{120,120,120}
\resizebox{0.95\linewidth}{!}{%
\begin{tikzpicture}[
  x=1cm, y=1cm,
  font=\scriptsize,
  >={Stealth[length=3.0pt,width=2.6pt]},
  cardS/.style={rounded corners=1.5pt, line width=0.5pt},
  stage/.style={text=white, font=\fontsize{6.6}{7.5}\selectfont\bfseries,
    inner sep=0pt},
  l1/.style={anchor=base, inner sep=0pt, text=black!85},
  l2/.style={anchor=base, inner sep=0pt, text=black!55,
    font=\fontsize{6.6}{7.5}\selectfont},
  qa/.style={anchor=base west, inner sep=0pt, text=black!75},
  dotS/.style={circle, inner sep=0pt, minimum size=0.11cm,
    line width=0.4pt},
  edge/.style={->, line width=0.45pt, dpLine}
]
\def\cw{1.42}\def\p{1.50}\def\hw{0.71}
\def\ch{0.19}\def\d{0.10}\def\g{0.06}
\pgfmathsetmacro{\W}{\p+\d-\g}  
\node[qa] at (0,0.44)
  {\textbf{Question:} which behaviors, at which rounds,
   does $\hat{y}$ depend on?};
\foreach \i/\name in {0/Observe,1/Measure,2/Hypothesize,
                      3/Test,4/Experiment}{
  \pgfmathsetmacro{\c}{\i*\p+\hw}
  \pgfmathsetmacro{\L}{\i==0 ? 0 : \c-\W/2}
  \pgfmathsetmacro{\R}{\i==4 ? \c+\hw : \c+\W/2}
  \pgfmathsetmacro{\n}{\i==0 ? 0 : \d}
  \fill[dpDark] (\L,\ch) -- (\R-\d,\ch) -- (\R,0)
    -- (\R-\d,-\ch) -- (\L,-\ch) -- (\L+\n,0) -- cycle;
  \pgfmathsetmacro{\tx}{(\L+\n/2+\R-\d/2)/2}
  \node[stage] at (\tx,0) {\strut\name};
}
\def\top{-0.30}\def\bot{-1.42}
\foreach \i/\fc/\dc in {0/white/black!45, 1/white/black!45,
    2/white/black!45, 3/dpGrey/dpGreyLine, 4/dpLight/dpLine}{
  \draw[cardS, fill=\fc, draw=\dc]
    (\i*\p,\top) rectangle (\i*\p+\cw,\bot);
}
\foreach \i/\a/\b in {
    0/Trajectories/{(CESR, \S4.1)},
    1/{Codes $\Omega$}/{(\S4.2)},
    2/{Parents $\hat{P}$}/{(GES, \S4.3)},
    3/{C1--C3 tests}/{(\S4.3)},
    4/{C4 steering}/{(\S5.2)}}{
  \node[l1] at (\i*\p+\hw,-1.00) {\a};
  \node[l2] at (\i*\p+\hw,-1.29) {\b};
}
\begin{scope}[shift={(\hw,-0.58)}]
  \draw[line width=0.45pt, black!45] (-0.48,0) -- (0.48,0);
  \node[dotS, fill=black!30, draw=black!45] at (-0.48,0) {};
  \foreach \x in {-0.24,0,0.24}
    \node[dotS, fill=dpLight, draw=dpLine] at (\x,0) {};
  \node[dotS, fill=dpDark, draw=dpDark] at (0.48,0) {};
\end{scope}
\begin{scope}[shift={(\p+\hw,-0.58)}]
  \foreach \r/\row in {0/{1,0,1,1,0,1},1/{0,1,0,1,1,0},
                       2/{1,1,0,0,1,0}}{
    \foreach \v [count=\k from 0] in \row{
      \pgfmathsetmacro{\xx}{-0.275+\k*0.11}
      \pgfmathsetmacro{\yy}{0.10-\r*0.10}
      \ifnum\v=1
        \fill[dpMid] (\xx-0.042,\yy-0.038) rectangle
                     (\xx+0.042,\yy+0.038);
      \else
        \draw[black!30, line width=0.3pt, fill=white]
          (\xx-0.042,\yy-0.038) rectangle (\xx+0.042,\yy+0.038);
      \fi
    }
  }
\end{scope}
\begin{scope}[shift={(2*\p+\hw,-0.58)}]
  \node[dotS, fill=white, draw=black!35] at (-0.40,0.14) {};
  \node[dotS, fill=dpLight, draw=dpLine] (a) at (-0.14,0.14) {};
  \node[dotS, fill=dpLight, draw=dpLine] (b) at ( 0.14,0.14) {};
  \node[dotS, fill=white, draw=black!35] at ( 0.40,0.14) {};
  \node[dotS, fill=dpDark, draw=dpDark] (y) at (0,-0.14) {};
  \foreach \s in {a,b}
    \draw[line width=0.45pt, dpLine, shorten <=0.4pt, shorten >=0.4pt,
      -{Stealth[length=2.3pt,width=2.1pt]}] (\s) -- (y);
\end{scope}
\begin{scope}[shift={(3*\p+\hw,-0.58)}]
  \foreach \k in {0,...,4}{
    \pgfmathsetmacro{\xx}{-0.40+\k*0.20}
    \ifnum\k=3
      \fill[black!55] (\xx-0.075,-0.11) rectangle (\xx+0.075,0.11);
    \else
      \draw[black!40, line width=0.35pt, fill=white]
        (\xx-0.075,-0.11) rectangle (\xx+0.075,0.11);
    \fi
  }
\end{scope}
\begin{scope}[shift={(4*\p+\hw,-0.58)}]
  \fill[black!35] (-0.20,-0.12) rectangle (-0.04,0.00);
  \fill[dpDark]   ( 0.04,-0.12) rectangle ( 0.20,0.12);
  \draw[line width=0.4pt, black!55] (-0.32,-0.12) -- (0.32,-0.12);
\end{scope}
\def\by{-1.56}
\foreach \a/\b/\t in {0/{2*\p+\cw}/hypothesis generation,
                      {3*\p}/{4*\p+\cw}/hypothesis testing}{
  \draw[line width=0.4pt, black!45]
    (\a,\by+0.06) -- (\a,\by) -- (\b,\by) -- (\b,\by+0.06);
  \node[l2, anchor=base] at ({(\a+\b)/2},\by-0.25) {\textit{\t}};
}
\node[qa] at (0,-2.13)
  {\textbf{Answer:} round-specific dependencies and
   evidence from C1--C4};
\end{tikzpicture}%
}
\caption{ReBeCA as a discovery pipeline. Coded trajectories
yield a hypothesized parent set $\hat{P}$ of $\hat{y}$, which
is assessed through C1--C3 (grey) on both tasks and
steering on unseen problems C4 (blue) for MATH.}
\label{fig:validation}
\end{figure}

%% file: Fig/cesr.tex
\begin{figure}[t]
\centering
\definecolor{cesrFb}{RGB}{238,238,238}
\definecolor{cesrFbLine}{RGB}{120,120,120}
\definecolor{cesrRf}{RGB}{214,230,246}
\definecolor{cesrRfLine}{RGB}{43,108,176}
\definecolor{cesrC1}{RGB}{33,90,160}
\definecolor{cesrC2}{RGB}{96,150,205}
\definecolor{cesrC3}{RGB}{150,190,225}
\definecolor{cesrC4}{RGB}{198,220,240}
\definecolor{cesrNoise}{RGB}{190,190,190}
\definecolor{cesrInk}{RGB}{15,30,60}
\resizebox{0.95\linewidth}{!}{%
\begin{tikzpicture}[
  x=1cm, y=1cm,
  font=\footnotesize,
  >={Stealth[length=3.6pt,width=3.2pt]},
  box/.style={draw=black!50, rounded corners=1.5pt,
    line width=0.5pt, minimum width=0.70cm,
    minimum height=0.48cm, inner sep=1pt, fill=white},
  fb/.style={box, fill=cesrFb, draw=cesrFbLine},
  rf/.style={box, fill=cesrRf, draw=cesrRfLine},
  flow/.style={->, line width=0.5pt, black!65},
  sq/.style={minimum size=0.19cm, inner sep=0pt,
    draw=white, line width=0.35pt},
  dot/.style={circle, minimum size=0.11cm, inner sep=0pt},
  head/.style={font=\scriptsize, text=black!75},
  lab/.style={font=\scriptsize, text=black!65}
]
\node[box] (x)  at (0.00,0) {$x$};
\node[rf]  (y0) at (0.95,0) {$y_0$};
\node[fb]  (f0) at (1.90,0) {$f_{b_0}$};
\node[rf]  (y1) at (2.85,0) {$y_1$};
\node[fb]  (f1) at (3.80,0) {$f_{b_1}$};
\node      (dd) at (4.62,0) {$\cdots$};
\node[rf]  (yT) at (5.45,0) {$y_T$};
\node[box] (yh) at (6.66,0) {$\hat{y}$};
\foreach \a/\b in {x/y0,y0/f0,f0/y1,y1/f1,f1/dd,dd/yT}
  \draw[flow] (\a) -- (\b);
\draw[flow, dashed] (yT) --
  node[above, lab, inner sep=1.5pt]{$g$} (yh);
\node[fb, minimum width=0.20cm, minimum height=0.17cm]
  (lf) at (4.90,0.50) {};
\node[lab, anchor=west, inner sep=1pt] at (lf.east) {feedback};
\node[rf, minimum width=0.20cm, minimum height=0.17cm]
  (lr) at (6.30,0.50) {};
\node[lab, anchor=west, inner sep=1pt] at (lr.east) {refine};
\draw[black!30, line width=0.4pt, rounded corners=3pt, fill=black!2]
  (-0.35,-0.78) rectangle (7.02,-3.10);
\draw[black!30, line width=0.4pt]
  (y1.south west) -- (-0.35,-0.78);
\draw[black!30, line width=0.4pt]
  (y1.south east) -- (7.02,-0.78);
\node[head] at (0.55,-1.18) {Sample ($K{=}20$)};
\node[head] at (3.45,-1.18) {Embed \& cluster};
\node[head] at (6.12,-1.18) {Select medoid};
\foreach \c [count=\i from 0] in
  {cesrC1,cesrC1,cesrC2,cesrC1,cesrC3,
   cesrC1,cesrNoise,cesrC1,cesrC2,cesrC4,
   cesrC3,cesrC1,cesrC2,cesrC1,cesrC4,
   cesrC2,cesrC1,cesrC3,cesrNoise,cesrC1}{
  \pgfmathtruncatemacro{\r}{int(\i/5)}
  \pgfmathtruncatemacro{\q}{mod(\i,5)}
  \node[sq, fill=\c] at ({0.05+0.25*\q},{-1.78-0.25*\r}) {};
}
\draw[flow] (1.30,-2.15) -- (2.05,-2.15);
\draw[cesrC1, line width=0.8pt]
  (3.20,-2.15) ellipse (0.50cm and 0.40cm);
\draw[cesrC2, dashed, line width=0.5pt]
  (4.28,-1.80) ellipse (0.30cm and 0.22cm);
\draw[cesrC3, dashed, line width=0.5pt]
  (4.30,-2.58) ellipse (0.28cm and 0.22cm);
\draw[cesrC4, dashed, line width=0.6pt]
  (2.55,-2.80) ellipse (0.22cm and 0.16cm);
\foreach \p in {(3.47,-2.12),(3.37,-1.93),(3.13,-1.87),(2.92,-1.99),
                (2.85,-2.21),(2.99,-2.38),(3.22,-2.43),(3.42,-2.33)}
  \node[dot, fill=cesrC1] at \p {};
\node[text=cesrInk, inner sep=0pt, font=\small]
  at (3.15,-2.15) {$\bigstar$};
\foreach \p in {(4.12,-1.77),(4.28,-1.68),(4.45,-1.84),(4.26,-1.92)}
  \node[dot, fill=cesrC2] at \p {};
\foreach \p in {(4.18,-2.56),(4.38,-2.48),(4.40,-2.68)}
  \node[dot, fill=cesrC3] at \p {};
\foreach \p in {(2.47,-2.82),(2.65,-2.77)}
  \node[dot, fill=cesrC4, draw=cesrC2,
    line width=0.3pt] at \p {};
\foreach \p in {(2.50,-1.62),(4.72,-2.86)}
  \node[dot, fill=cesrNoise] at \p {};
\draw[flow] (4.92,-2.15) -- (5.55,-2.15);
\node[rf, minimum width=0.80cm, minimum height=0.55cm]
  (out) at (6.12,-2.15) {$y_1$};
\node[text=cesrInk, inner sep=0pt, font=\scriptsize]
  at ([xshift=-1.5pt,yshift=-1.5pt]out.north east) {$\bigstar$};
\end{tikzpicture}%
}
\caption{Consistency-Enhanced Self-Refine (CESR). At each
step, $K = 20$ samples are embedded and clustered, and the
medoid ($\bigstar$) of the largest cluster becomes the
step output, so that each step reflects the model's typical output
rather than a single sample.}
\label{fig:second}
\end{figure}

%% file: Fig/graphs.tex
\begin{figure*}[t]
\centering
\definecolor{dpDark}{RGB}{33,90,160}
\definecolor{dpMid}{RGB}{96,150,205}
\begin{tikzpicture}[x=1cm,y=1cm,
  font=\fontsize{7}{8.4}\selectfont,
  cd/.style={circle,inner sep=0pt,minimum size=0.18cm,fill=white,draw=black!55,line width=0.5pt},
  par/.style={circle,inner sep=0pt,minimum size=0.2cm,fill=dpDark,draw=dpDark,line width=0.5pt},
  ring/.style={circle,inner sep=0pt,minimum size=0.33cm,fill=white,draw=black!50,line width=0.45pt,dash pattern=on 1.2pt off 0.8pt},
  yh/.style={circle,inner sep=0pt,minimum size=0.4cm,fill=dpDark,text=white,font=\fontsize{7.5}{8}\selectfont},
  chn/.style={-{Stealth[length=3pt,width=2.6pt]},dpMid,shorten <=0.8pt,shorten >=0.8pt},
  yed/.style={-{Stealth[length=3.4pt,width=3pt]},dpDark,line width=1.05pt,shorten <=0.8pt,shorten >=0.6pt},
  cnt/.style={font=\fontsize{6}{7}\selectfont,text=dpDark,inner sep=1pt},
  rnd/.style={text=black!55,font=\fontsize{6.5}{7.5}\selectfont},
  ttl/.style={font=\fontsize{7.5}{9}\selectfont\bfseries},
  rl/.style={anchor=east,text=black!80},
  nc/.style={text=black!45,font=\fontsize{6.5}{7.5}\selectfont\itshape},
  lg/.style={anchor=west,inner sep=1pt,text=black!80,font=\fontsize{6.6}{8}\selectfont}
]
\node[rl] at (-0.3,0) {Specificity};
\node[rl] at (-0.3,-0.64) {Clarity and Organization};
\node[rl] at (-0.3,-1.28) {Completeness};
\node[rnd,anchor=east] at (-0.3,0.33) {Round};
\node[ttl,anchor=base west] at (-0.1,0.62) {(a) MATH};
\node[rnd] at (0,0.33) {1};
\node[rnd] at (0.92,0.33) {2};
\node[rnd] at (1.84,0.33) {3};
\node[rnd] at (2.76,0.33) {4};
\node[rnd] at (3.68,0.33) {5};
\draw[line width=1.2pt,black!46] (0,0) -- (0.92,0);
\draw[line width=0.911pt,black!36] (0.92,0) -- (1.84,0);
\draw[line width=0.975pt,black!38] (0.92,-0.64) -- (1.84,-0.64);
\draw[line width=1.2pt,black!46] (1.84,-0.64) -- (2.76,-0.64);
\draw[line width=1.2pt,black!46] (2.76,-0.64) -- (3.68,-0.64);
\draw[line width=1.2pt,black!46] (0,-1.28) -- (0.92,-1.28);
\draw[line width=1.2pt,black!46] (0.92,-1.28) -- (1.84,-1.28);
\draw[line width=1.2pt,black!46] (1.84,-1.28) -- (2.76,-1.28);
\draw[line width=1.2pt,black!46] (2.76,-1.28) -- (3.68,-1.28);
\node[cd] (MATHSp1) at (0,0) {};
\node[cd] (MATHSp2) at (0.92,0) {};
\node[cd] (MATHSp3) at (1.84,0) {};
\node[cd] (MATHSp4) at (2.76,0) {};
\node[cd] (MATHSp5) at (3.68,0) {};
\node[cd] (MATHCO1) at (0,-0.64) {};
\node[par] (MATHCO2) at (0.92,-0.64) {};
\node[cd] (MATHCO3) at (1.84,-0.64) {};
\node[cd] (MATHCO4) at (2.76,-0.64) {};
\node[cd] (MATHCO5) at (3.68,-0.64) {};
\node[cd] (MATHCm1) at (0,-1.28) {};
\node[cd] (MATHCm2) at (0.92,-1.28) {};
\node[cd] (MATHCm3) at (1.84,-1.28) {};
\node[cd] (MATHCm4) at (2.76,-1.28) {};
\node[cd] (MATHCm5) at (3.68,-1.28) {};
\node[yh] (MATHY) at (4.46,-0.96) {$\hat{y}$};
\draw[chn,line width=1.1pt] (MATHCO1) -- node[cnt,below] {28/28} (MATHCO2);
\draw[chn,line width=0.779pt] (MATHSp1) -- node[cnt,above right] {16/28} (MATHCO2);
\draw[yed] (MATHCO2) to[out=-75,in=180] (1.24,-0.96) -- node[cnt,below,pos=0.72] {28/28} (4.04,-0.96) to[out=0,in=180] (MATHY);
\node[ttl,anchor=base west] at (5.58,0.62) {(b) BOUQuET};
\node[rnd] at (5.68,0.33) {1};
\node[rnd] at (6.6,0.33) {2};
\node[rnd] at (7.52,0.33) {3};
\node[rnd] at (8.44,0.33) {4};
\node[rnd] at (9.36,0.33) {5};
\draw[line width=1.155pt,black!44] (5.68,0) -- (6.6,0);
\draw[line width=1.2pt,black!46] (6.6,0) -- (7.52,0);
\draw[line width=1.2pt,black!46] (7.52,0) -- (8.44,0);
\draw[line width=1.2pt,black!46] (8.44,0) -- (9.36,0);
\draw[line width=1.11pt,black!43] (5.68,-0.64) -- (6.6,-0.64);
\draw[line width=1.2pt,black!46] (7.52,-0.64) -- (8.44,-0.64);
\draw[line width=1.2pt,black!46] (8.44,-0.64) -- (9.36,-0.64);
\draw[line width=1.065pt,black!41] (5.68,-1.28) -- (6.6,-1.28);
\draw[line width=0.93pt,black!37] (6.6,-1.28) -- (7.52,-1.28);
\draw[line width=0.795pt,black!32] (7.52,-1.28) -- (8.44,-1.28);
\node[ring] at (5.68,0) {};
\node[cd] (BQSp1) at (5.68,0) {};
\node[ring] at (6.6,0) {};
\node[cd] (BQSp2) at (6.6,0) {};
\node[ring] at (7.52,0) {};
\node[cd] (BQSp3) at (7.52,0) {};
\node[ring] at (8.44,0) {};
\node[cd] (BQSp4) at (8.44,0) {};
\node[ring] at (9.36,0) {};
\node[cd] (BQSp5) at (9.36,0) {};
\node[ring] at (5.68,-0.64) {};
\node[cd] (BQCO1) at (5.68,-0.64) {};
\node[ring] at (6.6,-0.64) {};
\node[cd] (BQCO2) at (6.6,-0.64) {};
\node[ring] at (7.52,-0.64) {};
\node[par] (BQCO3) at (7.52,-0.64) {};
\node[ring] at (8.44,-0.64) {};
\node[cd] (BQCO4) at (8.44,-0.64) {};
\node[ring] at (9.36,-0.64) {};
\node[cd] (BQCO5) at (9.36,-0.64) {};
\node[cd] (BQCm1) at (5.68,-1.28) {};
\node[cd] (BQCm2) at (6.6,-1.28) {};
\node[cd] (BQCm3) at (7.52,-1.28) {};
\node[ring] at (8.44,-1.28) {};
\node[cd] (BQCm4) at (8.44,-1.28) {};
\node[cd] (BQCm5) at (9.36,-1.28) {};
\node[yh] (BQY) at (10.14,-0.96) {$\hat{y}$};
\draw[chn,line width=1.1pt] (BQCO2) -- node[cnt,below,pos=0.4] {20/20} (BQCO3);
\draw[chn,line width=0.912pt] (BQCO1) to[bend left=28] node[cnt,above=0.5pt,pos=0.42] {15/20} (BQCO3);
\draw[chn,line width=0.725pt] (BQCm2) -- node[cnt,below right] {10/20} (BQCO3);
\draw[yed] (BQCO3) to[out=-75,in=180] (7.84,-1.02) -- (9.72,-1.02) to[out=0,in=187] (BQY);
\draw[yed] (BQCO4) to[out=-32,in=162] (BQY);
\node[par] (l0) at (-3.96,-1.86) {};\node[lg] (t1) at ([xshift=2pt]l0.east) {outcome neighbor in all runs};\draw[yed] ([xshift=8pt]t1.east) -- ++(0.42,0) node[lg] (t2) {edge into $\hat{y}$};\draw[chn,line width=0.9pt] ([xshift=8pt]t2.east) -- ++(0.42,0) node[lg] (t3) {edge into highlighted node};\draw[line width=0.9pt,black!40] ([xshift=8pt]t3.east) -- ++(0.42,0) node[lg] (t4) {same code in consecutive rounds};\node[ring] (r5) at ([xshift=13pt]t4.east) {};\node[cd] at (r5) {};\node[lg] at ([xshift=1pt]r5.east) {marginally associated with $\hat{y}$};
\end{tikzpicture}
\caption{Dependency structures learned by ReBeCA across $4F$ GES runs on both tasks. Edges shown appear in at least half of the runs (labels: run counts; 28 MATH runs, 20 per BOUQuET metric). This recurrence summary differs from CVLL-based set selection in Table~\ref{tab:selection}.}
\label{fig:structure}
\end{figure*}

%% file: Tab/selection.tex
\begin{table*}[t]
\centering
\small
\renewcommand{\arraystretch}{1.08}
\newcommand{\uv}[1]{\textcolor{black!50}{#1}}
\begin{minipage}[t]{0.43\textwidth}
\setlength{\tabcolsep}{5pt}
\begin{tabular*}{\linewidth}[t]{@{\extracolsep{\fill}}lc@{}}
\toprule
\textbf{(a) MATH} & \textbf{CVLL}$^{\uparrow}$ \\
\cmidrule(l){2-2}
Candidate set & Correctness \\
\midrule
$\boldsymbol{\{\mathrm{CO}_2, \mathrm{Sp}_4\}}$ & $\mathbf{-0.434{\pm}0.563}$ \\
$\{\mathrm{CO}_2\}$ & $-0.445{\pm}0.548$ \\
$\{\mathrm{CO}_2, \uv{\mathrm{MI}_2}\}$ & $-0.447{\pm}0.561$ \\
$\{\mathrm{CO}_2, \mathrm{Sp}_3, \uv{\mathrm{MI}_2}\}$ & $-0.449{\pm}0.608$ \\
$\{\mathrm{CO}_2, \mathrm{CO}_3, \mathrm{Sp}_2\}$ & $-0.457{\pm}0.688$ \\
$\{\mathrm{CO}_2, \mathrm{CO}_4, \mathrm{Sp}_3\}$ & $-0.464{\pm}0.726$ \\
$\{\mathrm{CO}_2, \uv{\mathrm{MI}_2}, \uv{\mathrm{PPL}_2}\}$ & $-0.466{\pm}0.616$ \\
$\{\mathrm{CO}_2, \mathrm{Sp}_4, \uv{\mathrm{MI}_4}\}$ & $-0.471{\pm}0.747$ \\
$\{\mathrm{CO}_2, \mathrm{DA}_5, \uv{\mathrm{PPL}_4}\}$ & $-0.481{\pm}0.667$ \\
$\{\mathrm{CO}_2, \uv{\mathrm{MI}_5}\}$ & $-0.482{\pm}0.680$ \\
$\{\mathrm{CO}_2, \mathrm{CO}_5, \uv{\mathrm{MI}_5}\}$ & $-0.532{\pm}0.849$ \\
$\{\mathrm{CO}_2, \uv{\mathrm{MI}_4}, \uv{\mathrm{PPL}_4}, \uv{\mathrm{PPL}_5}\}$ & $-0.602{\pm}1.105$ \\
\midrule
Null ($\emptyset$) & $-0.559{\pm}0.486$ \\
Full ($\Omega$, 30 variables) & $-0.861{\pm}1.227$ \\
\bottomrule
\end{tabular*}
\end{minipage}%
\hfill
\begin{minipage}[t]{0.55\textwidth}
\setlength{\tabcolsep}{3pt}
\begin{tabular*}{\linewidth}[t]{@{\extracolsep{\fill}}lcc@{}}
\toprule
\textbf{(b) BOUQuET} & \multicolumn{2}{c}{\textbf{CVLL}$^{\uparrow}$} \\
\cmidrule(l){2-3}
Candidate set & COMET & MetricX \\
\midrule
$\boldsymbol{\{\mathrm{CO}_3\}}$ & $\mathbf{-0.978{\pm}0.495}$ & $\mathbf{-1.003{\pm}0.536}$ \\
$\{\mathrm{CO}_3, \mathrm{CO}_4\}$ & $-0.981{\pm}0.736$ & $-1.007{\pm}0.733$ \\
$\{\mathrm{CO}_3, \mathrm{CO}_4, \uv{\mathrm{PPL}_5}\}$ & $-1.009{\pm}0.880$ & $-1.085{\pm}0.935$ \\
$\{\mathrm{CO}_3, \mathrm{CO}_4, \mathrm{Sp}_4\}$ & $-1.036{\pm}0.945$ & $-1.054{\pm}0.907$ \\
$\{\mathrm{CO}_3, \mathrm{CO}_4, \uv{\mathrm{PPL}_4}\}$ & $-1.088{\pm}0.976$ & $-1.075{\pm}0.960$ \\
$\{\mathrm{CO}_2, \mathrm{CO}_3, \mathrm{CO}_4, \uv{\mathrm{PPL}_3}\}$ & $-1.231{\pm}1.374$ & $-1.216{\pm}1.283$ \\
\midrule
Null ($\emptyset$) & $-1.044{\pm}0.337$ & $-1.044{\pm}0.342$ \\
Full ($\Omega$, 30 variables) & $-1.152{\pm}0.784$ & $-1.141{\pm}0.763$ \\
\bottomrule
\end{tabular*}

\vspace{9.3pt}
\setlength{\tabcolsep}{3pt}
\begin{tabular*}{\linewidth}{@{\extracolsep{\fill}}llll@{}}
\toprule
CO & Clarity and Organization & $X_t$ & code $X$ at round $t$ \\
Sp & Specificity & Bold & selected set \\
DA & Depth of Analysis & \uv{Grey} & uncertainty indicator \\
\uv{MI} & mutual information & \uv{PPL} & perplexity \\
\bottomrule
\end{tabular*}
\end{minipage}
\caption{Stages 1--2: candidate parent sets of $\hat{y}$ from GES across folds and scores, ranked by CVLL, i.e., how well each set predicts the outcome of held-out trajectories (mean CVLL across folds $\pm$ SD of individual held-out log probabilities). Bold marks the highest mean CVLL among candidates; selected sets also outperform the null and full models.}
\label{tab:selection}
\end{table*}

%% file: Tab/stability.tex
\begin{table}[t]
\centering
\small
\renewcommand{\arraystretch}{1.05}
\setlength{\tabcolsep}{2.4pt}
\begin{tabular*}{\columnwidth}{@{\extracolsep{\fill}}lcccc@{}}
\toprule
\multicolumn{5}{@{}l}{\textbf{(a) MATH: effect size and consistency (Stage 3)}} \\
 & & \multicolumn{2}{c}{Fold-group test ($p$)} & \\
\cmidrule(lr){3-4}
 & Coefficient [95\% CI] & Mean & Variance & $\eta^2_\mathrm{p}$$^{\downarrow}$ \\
\midrule
$\mathrm{CO}_2$ & $.298$ [$.259$, $.338$] & $.956$ & $.950$ & $.019$ \\
$\mathrm{Sp}_4$ & $.160$ [$.122$, $.198$] & $.860$ & $.317$ & $.045$ \\
Intercept & $.514$ [$.488$, $.540$] & $.942$ & $.571$ & $.024$ \\
\bottomrule
\end{tabular*}

\vspace{6pt}
\setlength{\tabcolsep}{2.6pt}
\begin{tabular*}{\columnwidth}{@{\extracolsep{\fill}}lcccccc@{}}
\toprule
\multicolumn{7}{@{}l}{\textbf{(b) BOUQuET: final versus initial translation (\%)}} \\
 & \multicolumn{3}{c}{COMET$^{\uparrow}$} & \multicolumn{3}{c}{MetricX$^{\downarrow}$} \\
\cmidrule(lr){2-4}\cmidrule(l){5-7}
$\mathrm{CO}_3$ coded & Better & Equal & Worse & Better & Equal & Worse \\
\midrule
No ($n{=}152$) & 24.3 & 38.8 & 36.8 & 36.2 & 39.5 & 24.3 \\
Yes ($n{=}248$) & 37.5 & 8.1 & 54.4 & 56.5 & 9.3 & 34.3 \\
\bottomrule
\end{tabular*}
\caption{Dependence on the selected parents. (a) MATH: linear probability coefficients (difference in success rate; mean and 95\% CI over $M = 20$ folds), compared across four fold groups in mean (ANOVA) and variance (Levene), with between-group effect sizes $\eta^2_\mathrm{p}$. (b) BOUQuET: final versus initial translation under each metric; Equal denotes equal metric scores; with $\mathrm{CO}_3$ coded, both Better and Worse outcomes are more frequent.}
\label{tab:stability}
\end{table}

%% file: Tab/examples.tex
\begin{table}[t]
\centering
\small
\renewcommand{\arraystretch}{1.1}
\setlength{\tabcolsep}{4pt}
\begin{tabular}{@{}p{0.29\columnwidth}p{0.29\columnwidth}p{0.29\columnwidth}@{}}
\toprule
\textbf{Reference} & \textbf{Round 2} & \textbf{Round 3} \\
\midrule
\textit{Please don't look at me like this.} & Please don't stare at me or look at me with disapproval. & Please don't look at me that way. \\
 & \textcolor{black!55}{COMET$^{\uparrow}$ .525} & \textcolor{black!55}{COMET$^{\uparrow}$ .784} \\
\midrule
\textit{Feeling guilty?} & Do you feel guilty? & Have you felt guilty lately? \\
 & \textcolor{black!55}{COMET$^{\uparrow}$ .417} & \textcolor{black!55}{COMET$^{\uparrow}$ .335} \\
\bottomrule
\end{tabular}
\caption{Two round-3 revisions coded as Clarity and Organization (Qwen3-8B on BOUQuET; Burmese and Greek sources): one removes content added in round 2 and raises COMET, the other adds content absent from the source and lowers it.}
\label{tab:qual}
\end{table}

%% file: Tab/steering.tex
\begin{table}[t]
\centering
\small
\setlength{\tabcolsep}{5pt}
\begin{tabular}{@{}lccccc@{}}
\toprule
 & \multicolumn{5}{c}{\textbf{Correct answers}$^{\uparrow}$} \\
\cmidrule(l){2-6}
\textbf{Steering} & \textbf{0.6B} & \textbf{1.7B} & \textbf{4B} & \textbf{8B} & \textbf{Total} \\
\midrule
None & 1 & 13 & 21 & 24 & 59 \\
$\mathrm{CO}_2$ & 1 & 14 & 26 & 23 & \textbf{64} \\
$\mathrm{Sp}_4$ & 1 & 10 & 33 & 18 & \textbf{62} \\
$\mathrm{CO}_2$ + $\mathrm{Sp}_4$ & 0 & 9 & 19 & 21 & 49 \\
\bottomrule
\end{tabular}
\caption{Steering either code alone raises correct solutions (59 $\to$ 64 or 62), whereas joint steering drops total to 49. Success rates differ significantly on Qwen3-4B (Holm-adjusted $p = .020$, Cochran's Q).}
\label{tab:optimization_results}
\end{table}

%% file: appendix.tex
\appendix

\newtcolorbox{promptbox}[1]{colback=black!2, colframe=black!40,
  boxrule=0.4pt, arc=1.5pt, left=3pt, right=3pt, top=2pt,
  bottom=2pt, fonttitle=\scriptsize\bfseries, coltitle=black,
  colbacktitle=black!8, title={#1}, fontupper=\scriptsize}
\newcommand{\bq}{\texttt{\textasciigrave\textasciigrave\textasciigrave}}
\section{Self-Refine}
\label{sec:isr}
Our case study is built upon trajectories produced by Self-Refine~\cite{madaan2023self}, a pioneering prompting pipeline for iterative self-reflection of LMs. Each trajectory is generated as a single multi-turn dialogue, initialized by the query $x$ and the initial response $y_0 \sim \mathcal{M}_\theta(x)$. Let $h_{t-1} = (x, y_0, f_{b_0}, \dots, f_{b_{t-2}}, y_{t-1})$ denote the dialogue history prior to round $t$, with base case $h_0 = (x, y_0)$. Each round $t \in \{1, \dots, T\}$ sequentially elicits critique feedback on $y_{t-1}$ followed by a revised response:
\begin{equation}
\begin{aligned}
f_{b_{t-1}} &= \mathcal{M}_\theta\left(h_{t-1} \,\|\, p_\text{fb}\right),\\
y_t &= \mathcal{M}_\theta\left(h_{t-1} \,\|\, p_\text{fb} \,\|\, f_{b_{t-1}} \,\|\, p_\text{refine}\right),
\end{aligned}
\label{eq:sr_appendix}
\end{equation}
where $\|$ denotes turn-level dialogue concatenation formatted via the chat template, and $p_\text{fb}$ and $p_\text{refine}$ are the task-specific prompts in Figure~\ref{fig:prompts}. In our case study, each generation step, including $y_0$, is realized by CESR (Section~\ref{sec:cesr}), which returns the medoid of the largest embedding-based cluster among $K$ samples rather than a single sample. Round $t$ is thus fully defined by the pair $(f_{b_{t-1}}, y_t)$, and every round-indexed variable in $\Omega$ (e.g., $\mathrm{CO}_t$) characterizes the behaviors manifest within this pair.

Issuing the same prompt pair at every round makes each round index a shared procedural position, comparable across queries (Section~\ref{sec:prob}): $\mathrm{CO}_2$ denotes the same stage in every trajectory, and, since every feedback turn requests all dimensions, a code's presence reflects what the generated reflection and revision act on rather than prompt variation. The steered conditions of C4 depart from this protocol by restricting reflection to the target dimension at the targeted round (Appendix~\ref{sec:steer}).

\begin{figure*}[t]
\centering
\begin{tcbraster}[raster columns=2, raster equal height=rows,
  raster column skip=0.02\textwidth, raster row skip=6pt,
  raster every box/.style={bottom=6pt}]
\begin{promptbox}{MATH: feedback prompt $p_\text{fb}$}
Please evaluate the [RESPONSE] across the following
dimensions:\par
1) **Content Accuracy**: Are there any factual errors or
misinterpretations?\par
2) **Completeness**: Does the [RESPONSE] address all parts
of the [QUERY]? Are key details or causes missing from the
[RESPONSE]?\par
3) **Specificity**: Are the causes specifically named rather
than broadly categorized?\par
4) **Clarity \& Organization**: Is the [RESPONSE] clearly
structured and easy to follow?\par
6) **Depth of Analysis**: Does the [RESPONSE] show
understanding beyond surface-level? Are connections between
different causes explained?
\end{promptbox}
\begin{promptbox}{BOUQuET: feedback prompt $p_\text{fb}$}
Please evaluate the [RESPONSE] (translation) across the
following dimensions:\par
1) **Translation Accuracy**: Is the meaning of the source
text accurately conveyed in the target language?\par
2) **Completeness**: Are all parts of the source text
translated? Is any information missing or omitted?\par
3) **Fluency**: Does the translation read naturally in the
target language? Are there any grammatical errors or awkward
expressions?\par
4) **Cultural Appropriateness**: Are cultural references and
context appropriately adapted for the target language?\par
5) **Consistency**: Are terminology and style consistent
throughout the translation?\par
6) **Clarity**: Is the translation clear and easy to
understand for native speakers of the target language?
\end{promptbox}
\begin{promptbox}{MATH: refinement prompt $p_\text{refine}$}
Please refine the previous [RESPONSE] to the [QUERY]
according to the [FEEDACK]. Ensure that:\par
- You address and make corresponding adjustments for each
piece of [FEEDACK].\par
- The content remains accurate and clearly expressed.\par
- Any missing information is supplemented.\par
- Any unclear expressions are improved.\par
Output a refined response to the [QUERY].
\end{promptbox}
\begin{promptbox}{BOUQuET: refinement prompt $p_\text{refine}$}
Please refine the previous translation [RESPONSE] according
to the [FEEDACK]. Ensure that:\par
- You address each piece of [FEEDACK] to improve translation
quality.\par
- The translation accurately conveys the meaning of the
source text.\par
- The target language flows naturally and is grammatically
correct.\par
- Cultural context is appropriately handled.\par
- Terminology and style remain consistent.\par
Output an improved translation of the source text.
\end{promptbox}
\end{tcbraster}
\caption{Self-Refine prompts, reproduced verbatim. The query
$x$ is the first user turn. The feedback prompt follows the
latest response and the refinement prompt follows the
generated feedback, both as user turns and identically at
every round.}
\label{fig:prompts}
\end{figure*}

\section{Reflective Codes}
\label{sec:sp}

\noindent\textbf{From expressions to behaviors.} While token-level wording varies across queries, reflective codes identify recurring conversational behaviors independent of surface phrasing. For example, ``I apologize for the mistake I made last night'' and ``Thank you for being so accommodating'' share almost no lexical overlap, yet both manifest the same polite interactive behavior. Surface expressions rarely recur across hundreds of trajectories, whereas conversational behaviors do; coding trajectories into discrete recurring behaviors provides stable variables for dependency estimation while remaining directly interpretable~\cite{zhong2025hicode}.

\noindent\textbf{Induction, consolidation, and codebook.} Supplementing Section~\ref{sec:struct}, candidate induction examined 30 randomly sampled intermediate trajectories per task, blinded to final outcomes ($y_T$ correctness or metric change). Task accuracy criteria in the feedback prompts (Figure~\ref{fig:prompts}, e.g., Content Accuracy in MATH or Translation Accuracy in BOUQuET) were excluded from $\Omega$ to reduce direct overlap between behavioral variables and the outcome definition. During consolidation, fine-grained stylistic feedback dimensions from translation (specifically Fluency and Consistency) frequently co-occurred in phrasing revisions and were merged into Clarity and Organization (CO). Cultural Appropriateness was too rare to retain under the prevalence threshold of Section~\ref{sec:struct} and was pruned. Table~\ref{tab:codebook} lists the four consolidated codes shared across both benchmarks.

A code is marked active ($1$) at round $t$ if either the feedback $f_{b_{t-1}}$ evaluates the dimension or the revision $y_t$ modifies the text along it, following the disjunctive wording in Figure~\ref{fig:annot_prompt}. On MATH, all four codes exceed the $1\%$ prevalence threshold across rounds, retaining all $4 \times 5 = 20$ active code variables (and $|\Omega| = 30$ when paired with $\mathrm{MI}_t$ and $\mathrm{PPL}_t$). On BOUQuET, code prevalence is $83.2\%$ of rounds for Completeness, $67.6\%$ for CO, and $43.5\%$ for Sp. Depth of Analysis (DA) never occurs ($0\%$), which is therefore constant on BOUQuET and cannot enter any dependency, leaving the $15$ active code variables of Section~\ref{sec:obs}.

\begin{table}[t]
\centering
\small
\renewcommand{\arraystretch}{1.1}
\begin{tabularx}{\columnwidth}{@{}
  >{\raggedright\arraybackslash}p{0.30\columnwidth}
  >{\raggedright\arraybackslash}X@{}}
\toprule
\textbf{Code} & \textbf{Dimension addressed by the feedback
or the revision} \\
\midrule
\multicolumn{2}{@{}l}{\textit{Shared by MATH and BOUQuET (MT)}} \\
\midrule
Completeness & coverage of all parts of the query or source
text, e.g., supplying omitted content \\
Specificity (Sp) & specificity of the content, e.g.,
replacing broad statements with specific ones \\
Clarity and Organization (CO) & structure and readability of
the response, e.g., reordering or clarifying steps \\
Depth of Analysis (DA) & reasoning beyond the surface level,
e.g., explaining connections between steps \\
\bottomrule
\end{tabularx}
\caption{Reflective codebook. All four codes are shared by
MATH and BOUQuET (MT).}
\label{tab:codebook}
\end{table}

\noindent\textbf{Annotation protocol.} GPT-5 annotates each code and round in an isolated call using the prompt template in Figure~\ref{fig:annot_prompt}. Each prompt supplies the query, reference output, previous response $y_{t-1}$, feedback $f_{b_{t-1}}$, and revision $y_t$, eliciting a binary judgment and a brief rationale. Figure~\ref{fig:annot_prompt} illustrates Completeness; prompts for the other three codes use the identical format, substituting the target dimension description in the Task Definition block with the corresponding clause in Table~\ref{tab:codebook}.

\noindent\textbf{Human agreement.}
For the human validation in Section~\ref{sec:struct}, three NLP researchers independently annotated 100 randomly sampled instances balanced across all four codes. GPT-5 annotations matched the human majority label on 89 of these 100 instances ($89\%$).

\begin{figure*}[p]
\centering
\begin{tcbraster}[raster columns=2, raster equal height=rows,
  raster column skip=0.02\textwidth,
  raster every box/.style={bottom=6pt}]
\begin{promptbox}{MATH: annotation prompt (Completeness)}
\#\# What to Know before Your Task\par
\#\#\# 1. Reflective Code\par
A **Reflective code** refers to a high-level
characterisation that summarises a kind of conversational
behaviours recognizable across multiple single-turn
dialogues generated by a language model (LM). For example,
the reflective code ``repeating'' could represent
repetitive wording within a single response from a LM. It is
a straightforward yet sensible way of quantifying and
structuring multi-turn dialogues. For instance, 5 responses
produced by a LM and from a multi-turn dialogue could be
quantified into a $5 \times N$ binary matrix by defining N
reflective codes, where the existence of a reflective code
$S_i$ ($i = 1, 2, \dots, N$) regarding response $k$ ($k = 1,
2, \dots, 5$) is denoted as **1** ($S_i$ appears somewhere
within the response $k$) or **0** ($S_i$ does NOT appear
anywhere within the response $k$).\par
We are conducting quantitative analyses on a batch of
LM-generated self-reflection data, represented in pure
texts, on the basis of math queries. Self-reflection denotes
a specific kind of iterative self-prompting framework,
which, for any input query, generates a multi-turn
self-reflect-and-refine dialogue to try reaching an optimal
response to the query. Analysing quantitatively this kind of
data is non-trivial, so we came out with **Reflective
code**. By treating each sample, i.e. a multi-turn
self-reflect-and-refine dialogue, especially comprising the
**feedback** of the self-reflection phases and the **refined
responses** based on the **feedback** and previous outputs,
as a binary reflective code matrix, we are able to quantify
the data and employ statistical means for our downstream
analyses.\par
\#\#\# 2. Notations\par
1. Text content wrapped in
\textless{}q\textgreater{}\textless{}/q\textgreater{} tags
represent a specific math problem.\par
2. Text content wrapped in
\textless{}y0\textgreater{}\textless{}/y0\textgreater{} tags
represent the language model's previous response to the math
problem.\par
3. Text content wrapped in
\textless{}y*\textgreater{}\textless{}/y*\textgreater{} tags
represent the ground truth or reference output to the math
problem.\par
4. Text content wrapped in
\textless{}fb\textgreater{}\textless{}/fb\textgreater{} tags
represent the revision feedback that the model generates
from self-reflecting on its previous response \bq{}y0\bq{}.\par
5. Text content wrapped in
\textless{}y1\textgreater{}\textless{}/y1\textgreater{} tags
represent the new, revised response that the model produces
by modifying its previous response \bq{}y0\bq{} based on the
revision feedback \bq{}fb\bq{}.\par
\#\# Task Definition\par
Please carefully read and understand the preceding
instructions and the meanings of the text variables
corresponding to each tag, and then complete the following
identification task: Evaluate and judge whether the model's
self-reflect-and-refine record below demonstrates an attempt
to optimize the completeness of the previous response, i.e.,
\bq{}fb\bq{} evaluates if all parts of the math problem were
solved and if there is any information missing or omitted,
or y1 makes the corresponding refinement from the
perspective of completeness.\par
\textless{}q\textgreater{}\{query\}\textless{}/q\textgreater{}\par
\textless{}y*\textgreater{}\{ground\_truth\}\textless{}/y*\textgreater{}\par
\textless{}y0\textgreater{}\{prev\_resp\}\textless{}/y0\textgreater{}\par
\textless{}fb\textgreater{}\{feedback\}\textless{}/fb\textgreater{}\par
\textless{}y1\textgreater{}\{refined\_resp\}\textless{}/y1\textgreater{}\par
Please think step-by-step carefully, making your judgement
by replying True (optimisation of **completeness** exists in
the provided record) or False (optimisation of
**completeness** NOT exist in the provided record), and
stating the reason for your judgement.
\end{promptbox}
\begin{promptbox}{BOUQuET: annotation prompt (Completeness)}
\#\# What to Know before Your Task\par
\#\#\# 1. Reflective Code\par
A **Reflective code** refers to a high-level
characterisation that summarises a kind of conversational
behaviours recognizable across multiple single-turn
dialogues generated by a language model (LM). For example,
the reflective code ``repeating'' could represent
repetitive wording within a single response from a LM. It is
a straightforward yet sensible way of quantifying and
structuring multi-turn dialogues. For instance, 5 responses
produced by a LM and from a multi-turn dialogue could be
quantified into a $5 \times N$ binary matrix by defining N
reflective codes, where the existence of a reflective code
$S_i$ ($i = 1, 2, \dots, N$) regarding response $k$ ($k = 1,
2, \dots, 5$) is denoted as \bq{}1\bq{} ($S_i$ appears
somewhere within the response $k$) or \bq{}0\bq{} ($S_i$
does NOT appear anywhere within the response $k$).\par
We are conducting quantitative analyses on a batch of
LM-generated self-reflection data, represented in pure
texts, on the basis of machine translation requests.
Self-reflection denotes a specific kind of iterative
self-prompting framework, which, for any input query,
generates a multi-turn self-reflect-and-refine dialogue to
try reaching an optimal response to the query. Analysing
quantitatively this kind of data is non-trivial, so we came
out with **reflective code**. By treating each sample, i.e.
a multi-turn self-reflect-and-refine dialogue, especially
comprising the **feedback** of the self-reflection phases
and the **refined responses** based on the **feedback** and
previous outputs, as a binary reflective code matrix, we
are able to quantify the textual data and employ statistical
means for our downstream analyses.\par
\#\#\# 2. Notations\par
1. Text content wrapped in
\textless{}q\textgreater{}\textless{}/q\textgreater{} tags
represents a specific translation request.\par
2. Text content wrapped in
\textless{}y0\textgreater{}\textless{}/y0\textgreater{} tags
represents the language model's previous response to the
request.\par
3. Text content wrapped in
\textless{}y*\textgreater{}\textless{}/y*\textgreater{} tags
represents the ground truth or reference translation.\par
4. Text content wrapped in
\textless{}fb\textgreater{}\textless{}/fb\textgreater{} tags
represents the revision feedback that the model generates
from self-reflecting on its previous response \bq{}y0\bq{}.\par
5. Text content wrapped in
\textless{}y1\textgreater{}\textless{}/y1\textgreater{} tags
represents the new, revised translation that the model
produces by modifying its previous response \bq{}y0\bq{}
based on the revision feedback \bq{}fb\bq{}.\par
\#\# Task Definition\par
Please carefully read and understand the preceding
instructions and the meanings of the text variables
corresponding to each tag, and then complete the following
identification task: Evaluate and judge whether the model's
self-reflect-and-refine record below demonstrates an attempt
to optimize the **completeness** of the previous
translation, i.e., \bq{}fb\bq{} evaluates if all parts of
the source text were translated and if there is any
information missing or omitted, or \bq{}y1\bq{} makes the
corresponding refinement from the perspective of
**completeness**.\par
\textless{}q\textgreater{}\{query\}\textless{}/q\textgreater{}\par
\textless{}y*\textgreater{}\{ground\_truth\}\textless{}/y*\textgreater{}\par
\textless{}y0\textgreater{}\{prev\_resp\}\textless{}/y0\textgreater{}\par
\textless{}fb\textgreater{}\{feedback\}\textless{}/fb\textgreater{}\par
\textless{}y1\textgreater{}\{refined\_resp\}\textless{}/y1\textgreater{}\par
Please think step-by-step carefully, making your judgement
by replying True (optimisation of **completeness** exists in
the provided record) or False (optimisation of
**completeness** NOT exist in the provided record), and
stating the reason for your judgement.
\end{promptbox}
\end{tcbraster}
\caption{Annotation prompts for the code Completeness. Prompts for the other codes use the same
template and differ in the target code. \{query\}, \{ground\_truth\}, \{prev\_resp\}, \{feedback\}, and
\{refined\_resp\} hold the query, the reference output,
$y_{t-1}$, $f_{b_{t-1}}$, and $y_t$ of the annotated round.}
\label{fig:annot_prompt}
\end{figure*}

\section{Structure Learning Details}

\subsection{Greedy Equivalence Search}
\label{sec:ges}
We use the GES estimator in pgmpy~\cite{ankan2024pgmpy}, following score-based greedy structure learning~\cite{chickering2002optimal}. It searches for score-improving graph changes subject to acyclicity and the constraints below:
\begin{equation}
\mathcal{G}_{k+1}=\mathop{\arg\max}_{\mathcal{G}'\in\mathcal{N}(\mathcal{G}_k)}
\mathcal{F}(D,\mathcal{G}'),
\label{eq:ges}
\end{equation}
where $\mathcal{N}(\mathcal{G}_k)$ denotes admissible local graph changes. We use AIC, BIC, BDeu, and BDs, a score-improvement tolerance of $10^{-6}$, and equivalent sample size $\alpha=10$ for the Bayesian scores.

\noindent\textbf{Prior constraints and temporal ordering.} Two structural constraints are strictly enforced: all outgoing edges from $\hat{y}$ are forbidden (since the task outcome is determined after all reflection rounds terminate), and all backward edges from any variable at round $t'$ to any variable at an earlier round $t$ ($t' > t$) are prohibited to preserve temporal precedence. Across $7 \times 4 = 28$ GES runs on MATH and $5 \times 4 = 20$ runs per BOUQuET outcome metric, no learned edge points from a later round to an earlier one.

\noindent\textbf{Methodological assumptions and boundaries.}
Selected parents represent direct predictive dependencies within the observed variable set $\Omega \cup \{\hat{y}\}$ (Section~\ref{sec:prob}) under the assumption that the distribution is Markov and faithful to a DAG consistent with round order. Causal sufficiency is not assumed; latent factors outside $\Omega$ (such as intrinsic query complexity or unmodeled token features) may act as unobserved common causes. ReBeCA uses the learned structure to nominate sparse sets of behavioral variables~\cite{chickering2002optimal}. We assess these hypotheses through held-out prediction and consistency checks (C1--C3), followed by prompt-based steering (C4).

\subsection{BDs Score and CVLL}
\label{sec:bds}
For a discrete outcome $\hat{y}$ with $r_{\hat{y}}$ levels and a candidate parent set $P$, the BDs local log marginal likelihood on dataset $D$~\cite{scutari2016empirical} is defined as:
\begin{equation}
\begin{aligned}
&\mathcal{L}_\text{BDs}(\hat{y} \mid P; D) =
\sum_{j: N_j > 0} \Bigg[ \log
\frac{\Gamma(\alpha/\tilde{q})}
     {\Gamma(\alpha/\tilde{q} + N_j)} \\
&\qquad + \sum_{k=1}^{r_{\hat{y}}} \log
\frac{\Gamma(\alpha/(\tilde{q} r_{\hat{y}}) + N_{jk})}
     {\Gamma(\alpha/(\tilde{q} r_{\hat{y}}))} \Bigg],
\end{aligned}
\label{eq:bds}
\end{equation}
where $N_{jk}$ denotes the count of instances in $D$ where $\hat{y} = k$ and the parent configuration $P$ takes its $j$-th state, with $N_j = \sum_k N_{jk}$. Standard BDeu uniformly partitions the equivalent sample size $\alpha$ across all $q = 2^{|P|}$ theoretical configurations. In sparse binary behavioral data, $q$ grows exponentially, diluting prior mass across unobserved states. BDs addresses sparse configuration coverage by allocating prior mass only over the $\tilde{q} = |\mathcal{J}^+|$ configurations actually observed in the sample ($N_j > 0$), providing robust score estimation for sparse discrete settings~\cite{scutari2018dirichlet}.

\noindent\textbf{Cross-validated log-likelihood (CVLL).}
CVLL averages held-out log probabilities obtained from conditional models fitted on the corresponding training folds:
\begin{equation}
\begin{aligned}
\mathrm{CVLL}(P)&=\frac{1}{F}\sum_{f=1}^{F}
\frac{\ell_f(P)}{|D_f^{\mathrm{te}}|},\\
\ell_f(P)&=\sum_{i\in D_f^{\mathrm{te}}}
\log\hat p_f(\hat y_i\mid P_i;D_f^{\mathrm{tr}}).
\end{aligned}
\label{eq:cvll}
\end{equation}
Here $D_f^{\mathrm{tr}}$ and $D_f^{\mathrm{te}}$ are the training and test folds; $\hat p_f$ is fitted using $D_f^{\mathrm{tr}}$, and $P_i$ is the candidate-variable configuration for test instance $i$. Table~\ref{tab:selection} reports the mean in Eq.~\eqref{eq:cvll} and the SD of individual held-out log probabilities. Values are negative as they represent logarithms of probabilities in $(0,1)$, where higher (less negative) scores indicate superior out-of-sample fit. Exponentiating CVLL yields the geometric mean probability assigned to observed outcomes, with equal fold weighting:
on MATH, the selected set achieves $\mathrm{CVLL} = -0.434$ ($e^{-0.434} \approx 64.8\%$), outperforming the marginal baseline ($\mathrm{CVLL} = -0.559$, $57.2\%$), uniform guessing ($\log 0.5 \approx -0.693$, $50.0\%$), and the full model ($\mathrm{CVLL} = -0.861$, $42.3\%$). Similarly on BOUQuET, the selected set achieves $\mathrm{CVLL} = -0.978$ (COMET, $e^{-0.978} \approx 37.6\%$) and $-1.003$ (MetricX, $36.7\%$), surpassing the marginal baseline ($-1.044$, $35.2\%$) and uniform guessing ($\log \frac{1}{3} \approx -1.099$, $33.3\%$).

\section{Case Study Configuration}
\label{sec:study_config}
Table~\ref{tab:config} summarizes the configuration of our empirical study.

\begin{table}[t] \centering \small \renewcommand{\arraystretch}{1.1} \begin{tabularx}{\columnwidth}{@{}   >{\raggedright\arraybackslash}p{0.30\columnwidth}   >{\raggedright\arraybackslash}X@{}} \toprule \textbf{Component} & \textbf{Setting} \\ \midrule Models & Qwen3-0.6B, 1.7B, 4B, 8B; 4-bit bitsandbytes checkpoints \\ Protocol & Self-Refine, $T = 5$ rounds (Figure~\ref{fig:prompts}) \\ CESR & $K = 20$; temperature $0.7$; top-p $0.8$; \texttt{text-embedding-3-large}; HDBSCAN, cosine distance \\ MATH & 105 problems (3 per level $\times$ topic cell); binary outcome; 417 trajectories \\ BOUQuET & 5 source languages into English; 100 items per model; ternary outcome under COMET or MetricX; 400 trajectories \\ Variables & $|\Omega| = 30$: 4 codes and 2 uncertainty indicators per round (Table~\ref{tab:variables}) \\ Annotation & GPT-5; human validation on 100 instances \\ Stages 1--2 & GES with AIC, BIC, BDeu, BDs; $F = 7$ (MATH), $F = 5$ (BOUQuET) \\ Stage 3 & $M = 20$ folds in 4 groups; ANOVA and Levene; $p > .05$ and $\eta^2_\mathrm{p} < .06$ \\ Steering & 50 AIME problems; 4 conditions; Cochran's Q and exact McNemar tests \\ \bottomrule \end{tabularx} \caption{Configuration of the case study.} \label{tab:config} \end{table}

\noindent\textbf{Models.}
The four dense Qwen3 models span 0.6B to 8B parameters within one model family~\cite{yang2025qwen3}. Pooling their trajectories supports behavioral analysis across these scales. All models run as 4-bit bitsandbytes-quantized checkpoints.

\noindent\textbf{MATH benchmark.}
We sample three problems from each of the 35 cells formed by crossing the five difficulty levels with the seven subjects in MATH~\cite{hendrycks2021math}, yielding 105 problems. Running all four models across these problems produces 417 complete trajectories (excluding three runs omitted due to incomplete generation). The outcome $\hat{y}$ is binary: 1 if the final answer extracted from $y_T$ matches the ground-truth reference, and 0 otherwise.

\noindent\textbf{BOUQuET benchmark.}
We evaluate five source languages from BOUQuET~\cite{andrews2025bouquet} (Bengali, Greek, Burmese, Western Persian, and Swahili) translated into English, crossing them with two granularities (sentence, paragraph) and two text domains (conversational, miscellaneous web text). Each model translates five items per cell across the resulting 20 cells, yielding 100 trajectories per model and 400 in total. Trajectories are evaluated using COMET~\cite{rei2020comet} and MetricX~\cite{juraska2023metricx}. Let $s(\cdot)$ denote the quality metric oriented such that higher values indicate superior translations (MetricX scores are negated). The outcome is categorized relative to the initial output $y_0$: Equal if $s(y_T)=s(y_0)$, Better if $s(y_T)>s(y_0)$, and Worse if $s(y_T)<s(y_0)$. Of 400 trajectories, 79 preserve the text verbatim; MetricX assigns equal scores to four additional trajectories with changed text.

\begin{table}[t]
\centering
\small
\renewcommand{\arraystretch}{1.1}
\begin{tabularx}{\columnwidth}{@{}
  >{\raggedright\arraybackslash}p{0.30\columnwidth}
  >{\raggedright\arraybackslash}X@{}}
\toprule
\textbf{Variable} & \textbf{Value 1 if \dots} \\
\midrule
$\mathrm{Completeness}_t$, $\mathrm{Sp}_t$, $\mathrm{CO}_t$, $\mathrm{DA}_t$ & the code is present at round $t$ (Table~\ref{tab:codebook}) \\
$\mathrm{PPL}_t$ & the perplexity of $y_t$ exceeds that of $y_{t-1}$ (base case $y_0$ for $t=1$) \\
$\mathrm{MI}_t$ & the mutual information estimate at round $t$ exceeds that at round $t-1$ \\
\midrule
$\hat{y}$ & MATH: correct vs.\ incorrect; BOUQuET: Better, Equal, or Worse metric score \\
\bottomrule
\end{tabularx}
\caption{Variables in $\Omega$ ($t = 1, \dots, 5$) and the task outcome.}
\label{tab:variables}
\end{table}

\noindent\textbf{Variable definition and binarization.}
Table~\ref{tab:variables} summarizes the 30 candidate variables in $\Omega$. Reflective codes are binary by definition. Continuous uncertainty metrics from \citet{yan2025entrospect}, namely perplexity (PPL) and mutual information (MI), are converted into discrete indicators based on whether their values increase relative to the preceding response ($y_t$ versus $y_{t-1}$, with $y_0$ as the base reference for $t=1$). This discretization keeps the full candidate space discrete for score evaluation in Appendix~\ref{sec:ges}.

\noindent\textbf{Fold partitioning.}
Stages 1--2 use leave-one-subject-out evaluation on MATH ($F=7$) and leave-one-source-language-out evaluation on BOUQuET ($F=5$). All model trajectories within the held-out subject or language are assigned to the test fold. Stage~3 separately uses $M=20$ disjoint random trajectory folds arranged into four equal groups to assess coefficient consistency. The effect-size criterion is the stricter of the two: with $F(3, 16)$, $\eta^2_\mathrm{p} < .06$ requires $F < 0.34$, whereas $p > .05$ requires only $F < 3.24$.

\noindent\textbf{Steering experiment (C4).}
\label{sec:steer}
The steering experiment evaluates whether modulating nominated behaviors alters self-reflection performance on 50 non-overlapping Olympiad-level problems from AIME~\cite{aime_1983_2024}. The 50 problems are held constant across all four models and four experimental conditions. Each trial executes CESR (Section~\ref{sec:cesr}) using the prompts detailed in Appendix~\ref{sec:isr}. Steered conditions apply only at the designated round: feedback and refinement prompts drop all general requirements to instruct the model to focus exclusively on the targeted behavior (Clarity and Organization at round 2; Specificity at round 4). All other rounds retain the standard prompts, and joint steering applies both modifications. Cochran's Q test~\cite{cochran1950comparison} evaluates the four matched conditions separately for each model, with Holm correction across four tests. Exploratory pairwise comparisons use two-sided exact McNemar tests, with Holm correction across all six condition pairs for each of four models (24 tests). Qwen3-4B yields $Q=12.853$ (raw $p=.00497$; adjusted $p=.01986$). For $\mathrm{Sp}_4$ versus joint steering, 15 problems favor $\mathrm{Sp}_4$ and one favors joint steering (raw $p=.000519$; adjusted $p=.01245$). No policy-versus-baseline comparison survives the same 24-test correction.

Qwen3-0.6B solves at most one problem across all conditions. Excluding Qwen3-0.6B, the remaining three models solve an average of 19.3 out of 50 problems in the unsteered baseline, 21.0 under $\mathrm{CO}_2$ steering, 20.3 under $\mathrm{Sp}_4$ steering, and 16.3 under joint steering. The lower aggregate performance under joint steering therefore persists when the smallest model is excluded.

\noindent\textbf{Computational resources.}
CESR draws 220 samples per trajectory (11 generation turns with $K = 20$ each), totaling approximately $1.8 \times 10^5$ generations across the 817 primary trajectories, with an embedding call per sample. Annotation requires one GPT-5 call per code and round (20 calls per trajectory, approximately $1.6 \times 10^4$ calls in total). The steering study adds 800 trajectories (50 problems, 4 models, 4 conditions), requiring a further 176,000 generations under CESR. Model inference was executed on two cloud-served NVIDIA RTX 4080 GPUs, while structure learning and statistical evaluations were conducted on CPU clusters.